%% file: main.tex
\documentclass[letterpaper,journal]{IEEEtran}

\input{macros}

\usepackage{relsize}
\usepackage{microtype}
\usepackage{makecell}

\newcommand\myvdots{\vdots}

\begin{document}

\title{HapticGiant: A Novel Very Large Kinesthetic\\Haptic Interface with Hierarchical Force Control}

\author{Michael Fennel, Markus Walker, Dominik Pikos, and Uwe D. Hanebeck,~\IEEEmembership{Fellow, IEEE}
  \thanks{This work was supported by the ROBDEKON project (13N16539) of the German Federal Ministry of Education and Research.}
  \thanks{All authors are with the Intelligent Sensor-Actuator-Systems Laboratory (ISAS), Institute for Anthropomatics and Robotics, Karlsruhe Institute of Technology (KIT), Germany
\{%
{\tt\footnotesize michael.fennel},
{\tt\footnotesize markus.walker},
{\tt\footnotesize dominik.pikos},
{\tt\footnotesize uwe.hanebeck}%
\}%
{\tt\footnotesize @kit.edu}
.}
}


\maketitle

\begin{abstract}
  Research in virtual reality and haptic technologies has consistently aimed to enhance immersion.
  While advanced head-mounted displays are now commercially available, kinesthetic haptic interfaces still face challenges such as limited workspaces, insufficient degrees of freedom, and kinematics not matching the human arm.
  In this paper, we present HapticGiant, a novel large-scale kinesthetic haptic interface designed to match the properties of the human arm as closely as possible and to facilitate natural user locomotion while providing full haptic feedback.
  The interface incorporates a novel admittance-type force control scheme, leveraging hierarchical optimization to render both arbitrary serial kinematic chains and Cartesian admittances.
  Notably, the proposed control scheme natively accounts for system limitations, including joint and Cartesian constraints, as well as singularities.
  Experimental results demonstrate the effectiveness of HapticGiant and its control scheme, paving the way for highly immersive virtual reality applications.
\end{abstract}


\section{Introduction}
\label{sec:into}

\IEEEPARstart{H}{ighly}
immersive virtual reality (VR) applications have been appearing in the media for decades.
Notable examples include the arcade game in the movie \emph{Tron} (1982), the \emph{Holodeck} from the \emph{Star Trek} series (1988), the whole story around \emph{Assassin's Creed} (2016), and the VR multiplayer game in \emph{Ready Player One} (2018).
As technology continues to evolve, the ideas that were long considered science fiction are now slowly becoming reality with the advent of consumer-grade augmented and virtual reality (AR/VR) headsets, including \emph{Oculus Rift} (2013), \emph{HTC Vive} (2016), \emph{Microsoft Hololens 2} (2019), and \emph{Apple Vision Pro} (2024).
With the most recent hardware, the visual aspects of VR are already covered so well that companies such as \emph{Meta} are betting on the technology's widespread success.
However, apart from simple vibrotactile feedback in smartphones and touch-displays, kinesthetic and tactile haptic feedback are still in their infancy, as evidenced by a growing variety of prototype devices~\cite{adilkhanovHapticDevicesWearabilityBased2022, kernEngineeringHapticDevices2023}.

\begin{figure}[t]
  \centering
  \subfigure[Proposed system with a user.]{
    \includegraphics[width=0.8\linewidth]{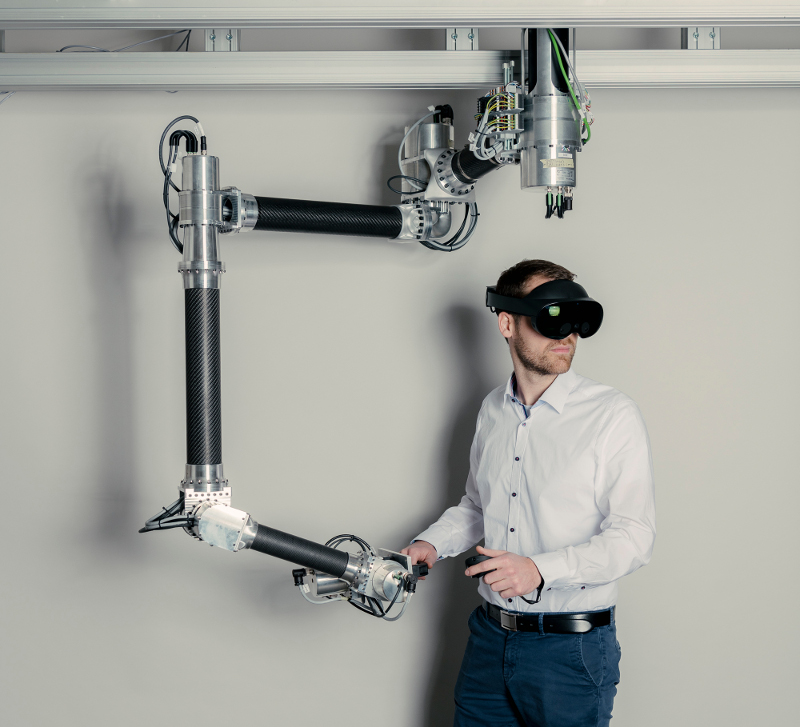}
    \label{fig:intro:system:user}
  }
  \subfigure[Visualization in \emph{rviz}, including measured joint torques, user wrench, and laser scanner data.]{
    \includegraphics[width=0.8\linewidth,trim={6cm 0 5cm 0cm},clip]{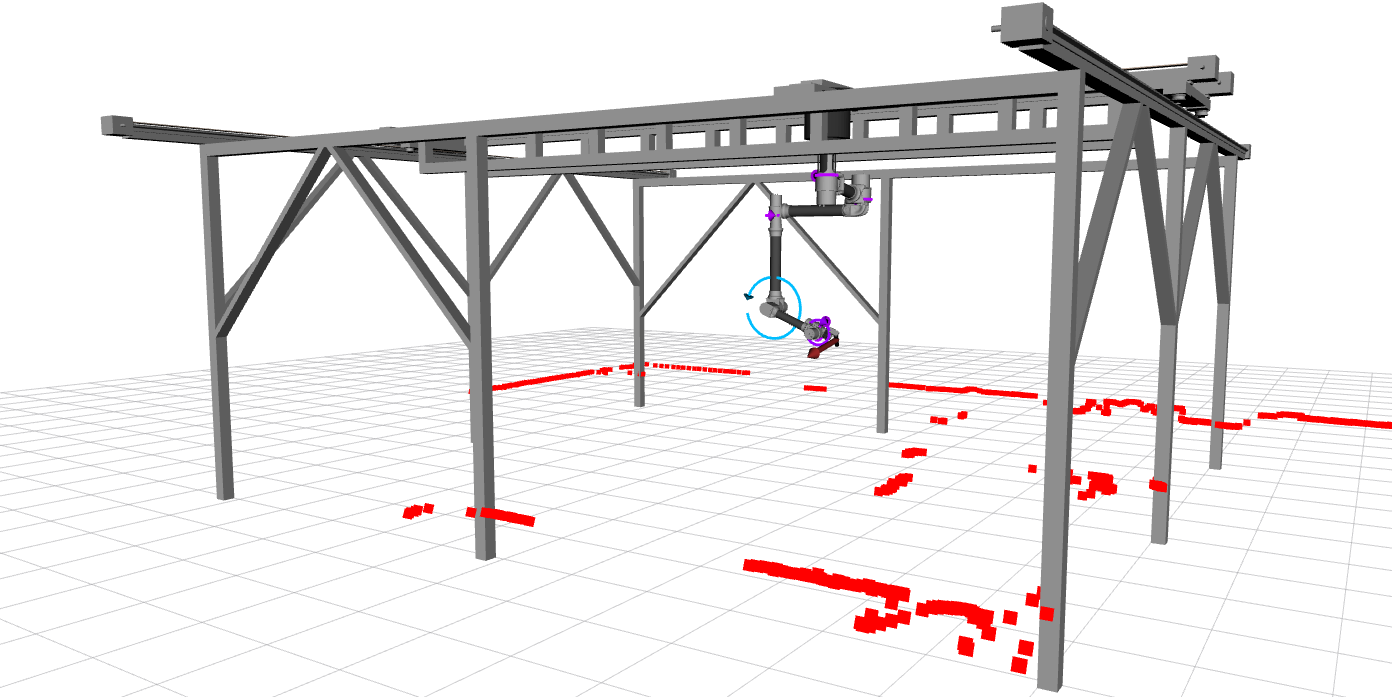}
    \label{fig:intro:system:visualization}
  }
  \caption{HapticGiant in reality and its 3D visualization.}
  \label{fig:intro:system}
\end{figure}

To deliver a truly immersive experience in the sense of the media mentioned above, a realistic interaction with objects in the target environment is required.
In this paper, we focus on the kinesthetic part of haptic feedback, which enables physical target environment interactions such as grasping, pushing, and pulling objects.
\subsection{Related Work}
The hardware for artificial kinesthetic sensations can be classified into wearable and grounded devices~\cite{hannafordHaptics2008}.
Wearable devices are typically gloves~\cite{caeiro-rodriguezSystematicReviewCommercial2021} or exoskeletons~\cite{shenChapterUpperLimb2020}.
A major drawback of this device category is that the user feels the generated forces at the attachment points of the device, reducing the achievable level of immersion. In addition, the advantage of a theoretically unlimited workspace comes at the cost of the user permanently carrying around a significant weight.
%

The grounded kinesthetic haptic interfaces from the database in~\cite{seifiHaptipediaAcceleratingHaptic2019} are usually designed as table-top devices, and thus, have a very limited workspace.
As demonstrated in the systems of~\cite{hulinDlrBimanualHaptic2011} and~\cite{lenzBimanualTeleop2023}, the workspace of the arms can be covered by grounded haptic devices. However, these systems remain stationary, thereby preventing the user from moving freely within a room-scale virtual environment. This substantially reduces the level of immersion that can be achieved because the user cannot move freely in the environment.

In addition to the classification outlined above, there are systems that attempt to overcome these workspace limitations by using different approaches. 
The \emph{Haption Inca 6D}~\cite{perretINCA6DCommercial2009} utilizes six motorized cables to control force feedback and spatial positioning of a handheld manipulator. This approach has the potential to extend the operational range of the haptic device; however, it imposes limitations on the user's mobility, due to the cables that traverse the workspace.
With the semi-mobile haptic interface~\cite{rosslerNovelHapticInterface2006b}, which was previously built in the authors' lab, the range of motion can be extended to the size of the portal frame. However, the system is unable to follow natural arm movements due to the lack of degrees of freedom. 
Some systems use a combination of grounded devices and a mobile platform to extend the workspace of the haptic interface. In~\cite{leeImprovementsMHI2009} and~\cite{dePascaleMobileHapticGrasping2006}, haptic table-top devices are mounted on omnidirectional mobile platforms. The system proposed in~\cite{peerMobileHapticForBimanualManipulation2008} incorporates a customized dual-arm configuration with force-feedback actuators on a mobile platform to enable haptic interaction in large virtual environments. However, the expansion of the workspace through the combination of grounded haptic devices and mobile platforms is associated with the disadvantage of complex positioning of the mobile platform for collision-free operation with the user.

While these systems are already quite advanced and have large workspaces, they still have one or more of the following limitations:
\begin{enumerate*}[label=(\roman*)]
  \item The haptic end effector has less than six degrees of freedom (DOF) and in some cases the rotational DOF are passive.
  \item Another restriction is that the workspace is still small compared to a human moving through a scene with natural locomotion.
  \item Even worse, the mechanical designs of the presented devices do not match the human arm at all.
  As a result, the achievable range of human arm motion is limited, resulting in weaker immersion.
\end{enumerate*}

On the control side, the presented interfaces usually focus on pure admittance or impedance control schemes.
This means that the consideration of hardware constraints such as joint limits, workspace boundaries, and singularities is not an integral part of the control scheme, but rather a post-processing step, if done at all.
The rendering of generalized, non-linear admittances with crisp motion limits as in~\cite{fennelHapticRenderingArbitrary2022} is also not integrated into the control scheme.
\subsection{Contribution}
To overcome these limitations, we introduce the novel, one-of-a-kind \emph{HapticGiant}, a grounded kinesthetic haptic interface, shown in \reffig{fig:intro:system}. In this work, our main contributions are as follows:
\begin{enumerate}
  \item We present a novel system design optimized to match the kinematic and dynamic properties of the human arm for good haptic transparency:
        The system is composed of a 2D gantry-crane-like prepositioning unit (PPU) linked to a custom six DOF manipulator.
        As a result, the system can achieve high forces and torques in any spatial direction within a room-scale workspace, allowing user locomotion through natural walking without interrupting the haptic rendering.
  \item To accurately render forces and torques, we propose a novel admittance-type force control scheme based on hierarchical optimization.
        This scheme facilitates the unified treatment of constraints such as joint limits, workspace limits, and singularity avoidance while the haptic rendering of arbitrary serial kinematic chains is active.
  \item Most importantly, a fully functional prototype is demonstrated.
        In combination with a suitable AR/VR visualization, this system is capable of generating a visual and haptic \emph{digital twin} (DT) of any serial kinematic chain, as illustrated in \reffig{fig:notation:definitions}.
\end{enumerate}

The structure of this paper is as follows.
Based on the notation in \refsec{sec:notation}, the system design, physical modeling, and software architecture are described in \refsec{sec:design}, \refsec{sec:modeling}, and \refsec{sec:architecture}, respectively.
The control scheme is introduced in \refsec{sec:control}.
Finally, experiments on a fully functional prototype are presented in \refsec{sec:experiments}.

\section{Coordinate Systems and Notation}
\label{sec:notation}

\begin{figure}
  \newcommand{\qppu}[1]{\q{\ppu}{[#1]}}
  \newcommand{\qadm}[1]{\q{\adm}{[#1]}}
  \newcommand{\qmanip}[1]{\q{\manip}{[#1]}}

  \definecolor{colorppu}{RGB}{0, 141, 36}
  \definecolor{colormanip}{RGB}{0, 133, 215}
  \definecolor{coloradm}{RGB}{206, 6, 6}

  \centering
  \def\svgwidth{\columnwidth}
  \footnotesize
  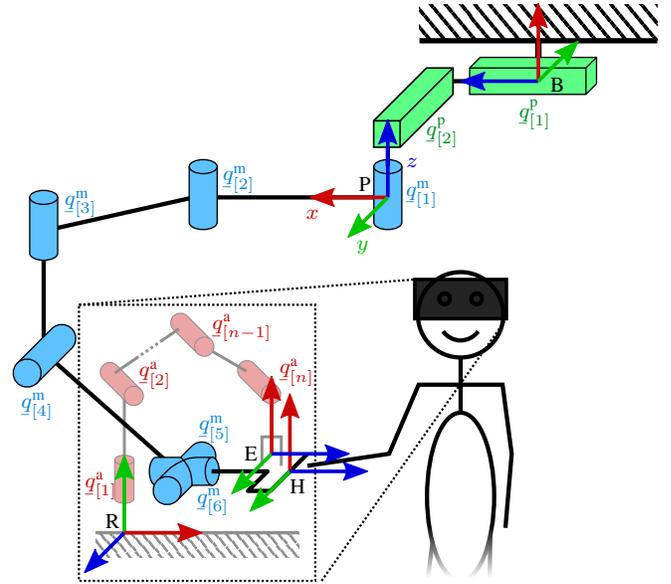
  \caption{
    Definition of the kinematic chains and coordinate systems used in this paper.
    The haptic interface is composed of the {\color{colorppu}PPU} and the {\color{colormanip}manipulator}.
    The {\color{coloradm}DT's} end effector frame E is kept coincident with H by the haptic rendering algorithm. The user observes the DT through a head-mounted display.}
  \label{fig:notation:definitions}
\end{figure}

Throughout this paper, postures including positions and translations are denoted by $\vek{x}$.
For orientations, rotation matrices $\mat{C}$ are used in conjunction with angular velocities $\vek{\omega}$.
Joint angles $\vek{q}$ and torques $\vek{\tau}$ are used to describe a kinematic chain in conjunction with the end effector Jacobian $\mat{J}$, the joint space inertia matrix $\mat{M}$, and the torque $\vek{c}$ summarizing all non-linear dynamic effects.
Stacking a pair of three-dimensional force and torque vectors results in a wrench $\vek{w} \in \mathbb{R}^6$.
The time derivative of a quantity is expressed with a dot, e.g., $\dot{\vek{x}}$.

Following the notation from~\cite{grovesPrinciplesGNSSInertial2013}, superscripts in Cartesian quantities denote the \emph{resolving coordinate frame}.
The subscripts denote the \emph{reference frame} (if applicable) and the \emph{object frame}, respectively.
For example, $\vek{x}{A}{BC}$ represents the position of point $C$ relative to frame $B$, given in coordinates of frame $A$.
Similarly, $\mat{C}{A}{B}$ describes the orientation of frame $B$ relative to frame $A$.
In the remainder, the frames shown in \reffig{fig:notation:definitions} are used.
Most importantly, B and H represent the base and the end effector, i.e., the handle, of the haptic manipulator, respectively.
Similarly, R and E denote the base and the end effector of the DT.

Non-Cartesian quantities referring to recurring entities such as the kinematic chain of the DT \enquote{\adm}, the PPU \enquote{p}, or the manipulator \enquote{m} are distinguished by the corresponding superscript.
For example, $\mat{J}{\manip}{}$ denotes the Jacobian of the manipulator's end effector.

To address specific elements of a vector or matrix, Matlab-style indexing is used in square brackets. For example, $\mat{A}{}{[1:3,:]}$ extracts the first three rows and all columns of the matrix $\mat{A}$.


\section{System Design}
\label{sec:design}

Most of HapticGiant was built from scratch, resulting in a unique design with outstanding capabilities.
This section presents the requirements, the design process, and some key aspects of the resulting hardware system.

\subsection{Requirements}
\label{sec:design:requirements}
The requirements that led to the overall design of HapticGiant can be categorized into three main areas: user experience, applications, and research.
Regarding the user experience, the system should most importantly provide good haptic transparency.
This implies that there are six Cartesian DOF, a large workspace, and good coverage of typical human hand poses and movements.
Furthermore, sufficient force and torque capabilities arise from the need to render stiff virtual environments.
As very close physical human-robot interaction is inherently necessary, a high degree of functional safety on various levels is required, including reliable emergency stop mechanisms.
On the application side, the system shall be available for tasks ranging in the spectrum from traditional teleoperation to fully simulated environments. In addition, the system shall be used in the context of motion compression~\cite{nitzscheMotionCompressionTelepresent2003}.
For research purposes, attributes such as extensibility, reconfigurability, repeatability, and low development effort are desired. This means that widespread tools in combination with as many off-the-shelf components as possible must be used to build HapticGiant.

\subsection{Dimensioning}
\label{sec:design:dimensioning}
Based on the above requirements and the experiences made with previous large kinesthetic haptic interfaces at the authors' lab~\cite{rosslerNovelHapticInterface2006b,perezariasHapticGuidanceExtended2013}, it was decided that an overhead-mounted manipulator on a PPU provides superior workspace properties.
During preliminary studies, it was found that a standard industrial six DOF manipulator is inadequate for the intended application, mainly due to unsuitable kinematics and limited access to the inner control workings.

For this reason, the optimization-driven design approach from~\cite{fennelOptimizationDrivenDesignKinesthetic2022a} was pursued.
The basic idea of this method is to analyze the capabilities of the human arm regarding reachable poses, velocities, and accelerations.
For this purpose, a seven DOF human arm model with kinematics, maximum joint angles, velocities, and accelerations is derived from anthropometric data.
Then, an analysis of the reachable workspace is performed.
For each reachable pose, an approximation of the reachable velocity and acceleration set is calculated using ellipsoid-shaped sets.
Based on practical experiences and the above requirements, the design problem is narrowed down to a specific kinematic structure with the decision variables being link lengths, joint sizes, and transmission ratios.
To convert this into an optimization problem, metrics are derived to quantify the relative coverage regarding pose, velocity, and acceleration as well as the force-torque capabilities of a manipulator design.
The resulting non-linear mixed-integer problem, which is enriched by constraints such as decision variable limits and monotonic joint sizes, is then solved using a genetic algorithm, yielding an optimal design for a manipulator with six DOF.

This manipulator design is ideal for a standing user right below the root joint of the six DOF manipulator
and provides a force of about \SI{50}{\newton} and a torque of about \SI{10}{\newton\meter} in any direction.
To account for the desired natural locomotion of the user, the manipulator is attached to a PPU with two orthogonal linear axes, whose size determines the walkable area. 
This combination of a six DOF manipulator and a PPU multiplies the optimized manipulator workspace to the entire area enclosed by the gantry structure. 
The PPU allows for large movements but high impedance, while the six DOF manipulator ensures accurate haptic rendering using series-elastic actuators.
Moreover, the utilization of this design approach ensures the haptic performance of the system remains consistent and independent of the location of the end effector. Using the hierarchical optimal control approach as proposed in~\refsec{sec:control}, the workspace of the haptic interface can be used efficiently.

\subsection{Construction}
\label{sec:design:construction}
A prototype system with eight DOF in a two-plus-six configuration was built based on the presented manipulator design.
The PPU measures $\SI{4.7}{\meter}$ times $\SI{5.5}{\meter}$ and is belt-driven by a total of four \emph{SEW Eurodrive} permanent-magnet synchronous motors with integrated mechanical brakes, all powered by matching \emph{Movidrive MDX61B} \citeDatasheet{SewMdx61b} inverters. \emph{Elgo EMAX} \citeDatasheet{ElgoEmax} absolute linear encoders provide position measurements.
The manipulator joint sizes, link lengths, and transmission ratios were taken directly from the optimized manipulator design, except that the transmission ratio of the last two joints was altered due to part availability and that the height of the first three joints was slightly adjusted for clearance reasons.
As a result, the manipulator is composed of six \emph{Sensodrive Sensojoint} \citeDatasheet{SensodriveSensojoint} series elastic actuators, whereas two \emph{7005}, two \emph{5005}, and two \emph{3008} actuators connect the links with a total link length of $\SI{2.5}{\meter}$ between the first actuator and the end effector.
Each Sensojoint is fully integrated, including an inverter, two absolute encoders, a safety-certified mechanical brake as well as a load torque sensor.
%

The mechanical construction was accomplished using the methods of system generation engineering~\cite{albersModelSGESystem2022}.
In particular, the use of carbon fiber tubes for the links was chosen to reduce the weight of the manipulator, while maintaining high stiffness.
A positive side effect of this choice is that cables can be routed mostly invisibly inside the tubes.
Some of the highly loaded aluminum parts, such as the flanges of joint 2, were created with the aid of topological optimization methods.
A custom \emph{B-Command rotarX} \citeDatasheet{BcommandRotarx} slip ring was integrated at the first joint, allowing infinite rotation of the manipulator as required for circular walking.
If the rotor of the slip ring is excluded, the total moving weight of the manipulator is $\SI{36,8}{\kilogram}$.
The replaceable end effector is a handle with an integrated dead-man switch, whose interaction forces and torques are captured using a \emph{Botasys SensONE} \citeDatasheet{BotasysSensone} six DOF force-torque sensor.
All sensors and actuators, which are not safety-relevant, are connected via the \emph{EtherCAT} fieldbus protocol to achieve low latency, high bandwidth, reduced effort for cabling, and less embedded development effort.

\subsection{Safety Concept}
By principle, a human user will be interacting with the manipulator's end effector and therefore reside in the security zone of the robot, which is prone to malfunction, especially when new algorithms are deployed.
Consequently, functional safety is a primary concern and guaranteed through a layered safety concept, that is independent of the actual controller running on HapticGiant:

\begin{enumerate} 
  \item At the lowest layer, the safety PLC level, two-channel signaling for safety-related sensors and actuators is used consistently.
        An override function for the dead-man switch at the end effector is available to allow autonomous operation without consent through grasping.
        In this case, a \emph{Sick TiM781S} \citeDatasheet{SickTim781S} safety laser scanner stops all movements if human presence is detected.
        The actual safety clearance is calculated by a \emph{Pilz PNOZ m B1} \citeDatasheet{PilzPnoz} safety PLC with \emph{PNOZ m EF 2MM} motion monitoring modules.

  \item To avoid undefined behavior in case of EtherCAT slave or communication failures, the EtherCAT watchdogs are utilized in the next layer of safety.

  \item In the software layer, a watchdog monitors whether system specifications, such as the permissible joint velocity or position range, are violated.
        A sanitizer module neutralizes setpoints causing the violation of system constraints, such as a positive torque command at an upper joint limit.
        Additionally, a collision detection algorithm based on \emph{FCL}~\cite{panFCLGeneralPurpose2012} predicts future system configurations based on the current velocity and halts the system before self-collisions can occur.

  \item The final safeguard is a human safety operator, who must press a separate dead-man switch to arm the system.
        Emergency stop buttons at various locations complete the safety concept.
\end{enumerate}

\section{Modelling}
\label{sec:modeling}

An accurate model of HapticGiant is essential for control and simulation purposes. For this reason, key aspects of kinematic and dynamic modeling are described in this chapter.

\subsection{Kinematics}
\label{sec:modeling:kinematics}
The kinematic structure of HapticGiant is illustrated in \reffig{fig:notation:definitions}.
The corresponding Denavit-Hartenberg (DH) parameters are derived from CAD data and given in \reftab{tab:problem:manipulator_dh}.
Together with simplified visual and collision geometry data, this is part of a URDF file describing the whole system.

Solving the inverse kinematics problem in closed form is possible.
For brevity, a detailed mathematical derivation is omitted here.
The basic idea is to formulate the end effector orientation using $zyz$-Euler angles.
Thus, the last joint angle $\q{\manip}{[6]}$ can be determined directly, and the sum $\q{\manip}{[4]} + \q{\manip}{[5]}$ is also known.
With the desired end effector height and basic trigonometry, $\q{\manip}{[4]}$ and $\q{\manip}{[5]}$ can be determined separately.
By interpreting the PPU position $\q{\ppu}$ as a parameter for describing redundancy, the remaining problem becomes planar and corresponds to a SCARA manipulator, whose solution is well-known from robotic fundamentals.

\begin{table}
  \centering
  \renewcommand*{\arraystretch}{1.2}
  \begin{tabular}{lcccc}
    \toprule
    Joint         & $\theta$                              & $d$                  & $a$                 & $\alpha$           \\ \midrule
    PPU 1         & 0                                     & ${\q{\ppu}{[1]}}$    & 0                   & $\SI{90}{\degree}$ \\
    PPU 2         & $\SI{90}{\degree}$                    & ${\q{\ppu}{[2]}}$    & 0                   & $\SI{90}{\degree}$ \\
    Manipulator 1 & $\q{\manip}{[1]}$                     & 0                    & $\SI{0.53}{\meter}$ & 0                  \\
    Manipulator 2 & $\q{\manip}{[2]}$                     & 0                    & $\SI{0.67}{\meter}$ & 0                  \\
    Manipulator 3 & $\q{\manip}{[3]} + \SI{180}{\degree}$ & $-\SI{0.75}{\meter}$ & 0                   & $\SI{90}{\degree}$ \\
    Manipulator 4 & $\q{\manip}{[4]} - \SI{90}{\degree}$  & 0                    & $\SI{0.42}{\meter}$ & 0                  \\
    Manipulator 5 & $\q{\manip}{[5]} + \SI{90}{\degree}$  & 0                    & 0                   & $\SI{90}{\degree}$ \\
    Manipulator 6 & $\q{\manip}{[6]}$                     & $\SI{0.12}{\meter}$  & 0                   & 0                  \\
    \bottomrule
  \end{tabular}
  \caption{DH parameters of HapticGiant.}
  \label{tab:problem:manipulator_dh}
\end{table}


\subsection{Dynamics}
\label{sec:modeling:dynamics}

A dynamic model of the system was obtained using the density distribution of the individual parts from the CAD model.
The weight of each part was updated with real measurements.
Additionally, the cable weights were measured and distributed along the links according to the link lengths and the respective cable routes.
The resulting model is used in control for feedforward compensation, especially for gravity compensation, and in simulation.

\section{Software Architecture}
\label{sec:architecture}

\begin{figure}
  \centering
  \def\svgwidth{\columnwidth}
  \footnotesize
  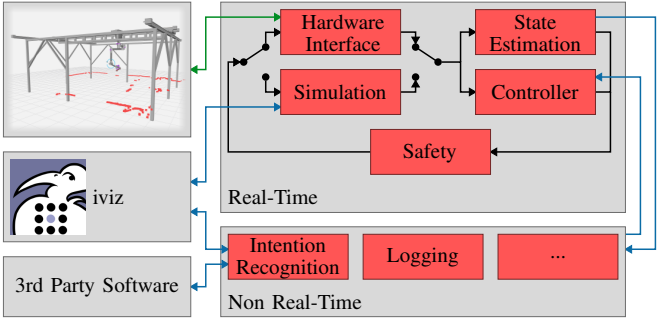

  \caption{Overview of the software architecture. All boxes on the right-hand side are running on standard computer hardware.}
  \label{fig:architecture:overview}
\end{figure}

Based on the requirements from \refsec{sec:design:requirements}, the architecture shown in \reffig{fig:architecture:overview} was designed.
For brevity, not all parts are covered in this paper, as emphasis is put on the overall system and the high-level control concept.
Specifically, we leave out a promising novel method for the kinematic state estimation, covered in-depth in~\cite{fennelCalibrationfreeIMUbasedKinematic2022a} and~\cite{fennelObservabilitybasedPlacementInertial2023}, and a method for user intention recognition~\cite{fennelIntentionEstimationRecurrent}.

\subsection{Real-Time Control Framework}
\label{sec:architecture:rtcf}

\emph{Robot Operating System} (ROS) is a widely used framework for robotic applications but lacks real-time capabilities in version~1.
When HapticGiant's software architecture was defined, the successor ROS~2 and its newly introduced real-time concept were immature.
To this day, ROS~2 still requires a lot of manual programming to achieve real-time performance.
For this reason, the \emph{Real-Time Control Framework} (RTCF)~\cite{fennelRTCFFrameworkSeamless2021a} was developed.
With a focus on modularity, interoperability, performance, and usability for developers coming from the ROS ecosystems, the system offers good real-time characteristics, low overhead, and full compatibility with ROS~1.
As a result, control frequencies far beyond $\SI{2}{\kilo\hertz}$ are easily possible if a well-tuned \emph{Preempt-RT} kernel is used.

\subsection{Simulation}
\label{sec:architecture:simulation}

As in most robotic applications, simulation is very beneficial for developing and testing new algorithms, as it reduces the risk of damage and increases the system's availability.
We found that existing tools either cannot run in real time, which is needed for live human-machine interaction, or that they cannot model all desired effects.
Thus, a custom simulation tool was developed, which can be used as a drop-in replacement for the real haptic interface in \reffig{fig:architecture:overview}.
The sensor models for encoders, inertial measurement units, as well as force and torque sensors include effects such as noise, bias, temperature drift, saturation, quantization, cross-coupling, and gain errors, where applicable.
The simulation of the mechanism is based on the dynamic model, with the addition that the PPU behavior is integrated using a hybrid dynamics formulation~\cite{featherstoneRigidBodyDynamics2008a} due to the black-box nature of the PPU drivetrain.
The manipulator joints are modeled as series elastic actuators to facilitate mechanical oscillations and the Coulomb friction is incorporated through the principle of maximum dissipation~\cite{drumwrightModelingContactFriction2011}.

\subsection{Visualization}
\label{sec:architecture:visualization}

The URDF model from \refsec{sec:modeling:kinematics} can be used for system visualization as depicted in \reffig{fig:intro:system:visualization}.
An appropriate visualization of the target environment is necessary to fully immerse users in a given application.
For this purpose, \emph{iviz}~\cite{zeaIvizROSVisualization2021a} is employed as the default visualization tool within the HapticGiant platform.
The iviz app offers functionality comparable to the ROS tool \emph{rviz} but is based on the \emph{Unity} game engine.
As a result, it enables effortless visualization of robotic systems across various AR/VR platforms, such as \emph{Apple iPad}, \emph{Microsoft Hololens}, and \emph{HTC Vive}.

\section{Control}
\label{sec:control}

Controlling HapticGiant is intricate because of multiple constraints and the need for real-time performance.
In the following, the proposed control architecture based on \emph{Hierarchical Quadratic Programming} (HQP) is presented.

\subsection{Hierarchical Quadratic Programming}
\label{sec:control:hqp}
\newcommand{\flb}[1]{\vek{f}_{\text{lb},#1}}
\newcommand{\fub}[1]{\vek{f}_{\text{ub},#1}}
\newcommand{\slackeq}[1]{\vek{w}{}{#1}}
\newcommand{\slackiq}[1]{\vek{v}{}{#1}}
\newcommand{\slackeqopt}[1]{\vek{w}{\ast}{#1}}
\newcommand{\slackiqopt}[1]{\vek{v}{\ast}{#1}}

The mathematical formulation in this section is based on~\cite{bellicosoPerceptionlessTerrainAdaptation2016a} and~\cite{herzogMomentumControlHierarchical2016}.
The basic idea of hierarchical optimal control is to define the control problem with the decision variable $\vek{x}$ as a set of tasks with priorities $p \in \mathbb{N}^+$.
Each task is of the form
\begin{equation}
  \label{eq:control:hqp:task}
  T_p :
  \begin{cases}
    \mat{A}_p \vek{x} + \slackeq{p} = \vek{b}_p \\
    \flb{i} \le  \mat{C}_p \vek{x} + \slackiq{p}  \le \fub{i}
  \end{cases}
  ,
\end{equation}
where $\slackeq{p}$ and $\slackiq{p}$ are slack variables of appropriate size.
The matrices $\mat{A}_p$, $\mat{C}_p$ and the vectors $\vek{b}_p$, $\flb{i}$, $\fub{i}$ are task-dependent.
Starting with $p=1$, the quadratic problem (QP)
\begin{equation}
  \label{eq:control:hqp:basic_qp}
  \begin{split}
    \left(\vek{x}^\ast_p, \vek{w}^\ast_p, \vek{v}^\ast_p \right) &=
    \underset{\vek{x}_p, \vek{w}_p, \vek{v}_p }{\mathrm{argmin}}
    \frac{1}{2}\norm{\vek{w}_p}^2 + \frac{1}{2}\norm{\vek{v}_p}^2 \\
    \text{subject to} \mathspace \mathspace
    & \mat{A}_1 \vek{x}_p  + \slackeqopt{1} = \vek{b}_1 \mathspace, \\
    & \myvdots \\
    & \mat{A}_{p-1} \vek{x}_{p}  + \slackeqopt{p-1} = \vek{b}_{p-1} \mathspace, \\
    & \mat{A}_{p} \vek{x}_{p}  + \slackeq{p} = \vek{b}_{p} \mathspace, \\
    & \flb{1} \le  \mat{C}_1 \vek{x}_p + \slackiqopt{1}  \le \fub{1} \mathspace, \\
    & \myvdots \\
    & \flb{p-1} \le  \mat{C}_{p-1} \vek{x}_p + \slackiqopt{p-1}  \le \fub{p-1} \mathspace, \\
    & \flb{p} \le  \mat{C}_{p} \vek{x}_p + \slackiq{p}  \le \fub{p} \mathspace
  \end{split}
\end{equation}
is solved iteratively.
Hence, lower priority tasks can find their solution within the set of equally good solutions of the higher priority tasks.
As the equality constraints of the tasks with priority $i < p$ must be fulfilled up to slack $\vek{w}^\ast_i$, the solution of the next priority is constrained to the null space of all the previous tasks' equality constraints.
Mathematically, this is expressed as
\begin{equation}
  \vek{x}_p = \vek{x}{\ast}{p-1} + \mat{Z}_{p} \vek{z}_p \mathspace,
  \label{eq:control:hqp:x_substitution}
\end{equation}
where
$\mat{Z}_{p} = \operatorname{kern}\!
  \left(
  \begin{pmatrix}
    \mat{A}{\tranT}{1} & \hdots & \mat{A}{\tranT}{p-1}
  \end{pmatrix}
  \!{}\tran
  \right)
$
with
$
  \mat{Z}_0 = \mat{I}
$
is the null space of all equality constraints merged before task $p$.
As shown in~\cite{bellicosoPerceptionlessTerrainAdaptation2016a}, the recursion
\begin{equation}
  \label{eq:control:hqp:kernel_recursion}
  \mat{Z}{}{p} = \mat{Z}{}{p-1} \operatorname{kern}(\mat{A}{}{p-1} \mat{Z}{}{p-1})
\end{equation}
improves the computational efficiency of the null space calculation.
The optimization problem \eqref{eq:control:hqp:basic_qp} is simplified by inserting the last equality constraint into the objective.
Additionally, the substitution from \eqref{eq:control:hqp:x_substitution} is applied to automatically fulfill the remaining equality constraints. This yields
\begin{equation}
  \begin{split}
    &\left(\vek{z}^\ast_p, \vek{v}^\ast_p \right)
    \!=\!
    \underset{\vek{z}_p, \vek{v}_p }{\operatorname{argmin}}
    \frac{1}{2}
    \!\norm{\mat{A}_p (\vek{x}{\ast}{p-1} \!+\! \mat{Z}_{p} \vek{z}_p)
      \!-\!
      \vek{b}_{p} }^2
    \!+\!
    \frac{1}{2}\!\norm{\vek{v}_p}^2 \\
    &\begin{split}
      \;\, \text{s. t.} \mathspace \mathspace
      & \flb{1} \le  \mat{C}_1 (\vek{x}{\ast}{p-1} +\mat{Z}_{p} \vek{z}_p) +\slackiqopt{1}  \le \fub{1} \mathspace, \\
      & \myvdots \\
      & \flb{p-1} \le  \mat{C}_{p-1} (\vek{x}{\ast}{p-1} + \mat{Z}_{p} \vek{z}_p)  + \slackiqopt{p-1}  \le \fub{p-1} \mathspace, \\
      & \flb{p} \le  \mat{C}_{p} (\vek{x}{\ast}{p-1} + \mat{Z}_{p} \vek{z}_p) + \slackiq{p}  \le \fub{p} \mathspace
    \end{split}
  \end{split}
\end{equation}
with
$\vek{x}{\ast}{p} = \vek{x}{\ast}{p-1} + \mat{Z}_{p} \vek{z}{\ast}{p}$
and
$\vek{x}{\ast}{0} = \vek{0}$,
which can be easily rewritten as a standard QP
of the form
\begin{equation}
  \begin{split}
    &\vek{\zeta}{\ast}{p} =
    \underset{\vek{\zeta}_p}{\operatorname{argmin}}
    \mathspace
    \frac{1}{2} \vek{\zeta}{\tranT} \mat{H}{}{p} \vek{\zeta} + \vek{g}{\tranT}{p} \vek{\zeta}
    \\
    &\text{subject to} \mathspace \mathspace
    \vek{\tilde{f}}{}{\text{lb},p} \le  \mat{\tilde{C}}{}{p} \vek{\zeta}_p \le \vek{\tilde{f}}{}{\text{ub},p} \mathspace,
  \end{split}
\end{equation}
where $ \vek{\zeta}{\tranT}{p} = \begin{pmatrix}\vek{z}{\tranT}{p} & \vek{v}{\tranT}{p}\end{pmatrix}$.
To avoid infeasible inequalities caused by numerical issues, small epsilons are added to the inequality bounds $\vek{\tilde{f}}{}{\text{lb},p}$ and $\vek{\tilde{f}}{}{\text{ub},p}$. Furthermore, the matrix $\mat{H}{}{p}$ is regularized by adding small epsilons to its diagonal.

\subsection{Joint Limit Handling}
\label{sec:control:limits}
Let $q$ and $\dot{q}$ be the joint position and velocity, respectively, of a joint with maximum ratings regarding position $q_\text{min}$ and $q_\text{max}$, velocity $\pm \dot{q}_\text{max}$, and acceleration $\pm \ddot{q}_\text{max}$.
If this joint state is under time-discrete control with sample time $\Delta t$ and acceleration input $\ddot{q}$, suitable joint acceleration limits can be calculated to avoid the violation of the joint ratings.
In~\cite{preteJointPositionVelocity2018}, the four acceleration intervals
\begin{equation}
  B\begin{cases}
    A\begin{cases}
      \phantom{[]} \\
      \phantom{[]}
    \end{cases} \\
    \phantom{[]}                    \\
    \phantom{[]}
  \end{cases}
  \kern-3.0em
  \begin{aligned}
    {[\ddot{q}_{\text{lb},1}, \ddot{q}_{\text{ub},1}]} & = \operatorname{accBoundsFromPosLimits}(q, \dot{q}) \mathspace,                                          \\
    {[\ddot{q}_{\text{lb},2}, \ddot{q}_{\text{ub},2}]} & = \nicefrac{1}{\Delta t} {[(-\dot{q}_\text{max} - \dot{q}), (\dot{q}_\text{max} - \dot{q})]} \mathspace, \\
    {[\ddot{q}_{\text{lb},3}, \ddot{q}_{\text{ub},3}]} & = \operatorname{accBoundsFromViability}(q, \dot{q}) \mathspace,                                          \\
    {[\ddot{q}_{\text{lb},4}, \ddot{q}_{\text{ub},4}]} & = {[-\ddot{q}_\text{max}, \ddot{q}_\text{max}]}
  \end{aligned}
  \kern-1em
\end{equation}
are derived for this purpose.
If the intersection of the first two intervals is taken as an admissible acceleration range, position and velocity limits are respected, but the acceleration may become arbitrarily large.
The resulting upper and lower bounds in vector notation are denoted with $\vek{l}{}{\text{lb},A}(\q{}, \dq{})$ and $\vek{l}{}{\text{ub},A}(\q{}, \dq{})$, respectively.
If the intersection of all intervals is considered, position, velocity, and acceleration limits are respected at all times, including the time between two samples.
For a kinematic chain, we call the resulting upper and lower bounds $\vek{l}{}{\text{lb},B}(\q{}, \dq{})$ and $\vek{l}{}{\text{ub},B}(\q{}, \dq{})$, respectively.

One core assumption in~\cite{preteJointPositionVelocity2018} is that the joint state never leaves the \emph{viable} set, which is unrealistic due to numerical errors in the integration step or tracking errors in practical scenarios.
Therefore, $\q{}$ is clamped to the position limits.
In a second step, $\dq{}$ is clamped according to the definition of the viable set (10) in~\cite{preteJointPositionVelocity2018}.
The problem here is that it cannot drive a physical joint position back to its feasible range once it has left the viable set because the clamping acts as if the position limits were never left.
Omitting the clamping is not an option, because non-viable states can yield empty acceleration intervals.
As a workaround, the acceleration interval
\begin{equation}
  {[
  \ddot{q}_{\text{lb},5}, \mathspace
  \ddot{q}_{\text{ub},5}
  ]}
  = \nicefrac{2}{\Delta t^2}
  {[
  {q}_{\text{min}} - q - \Delta t \dot{q}, \mathspace
  {q}_{\text{max}} - q - \Delta t \dot{q}
  ]} \mathspace,
\end{equation}
based on the second-order Taylor approximation of $q(t)$ as found in~\cite{bellicosoPerceptionlessTerrainAdaptation2016a}, is proposed.
The corresponding vector notation is defined as $\vek{l}{}{\text{lb},C}(\q{}, \dq{})$ and $\vek{l}{}{\text{ub},C}(\q{}, \dq{})$.
While velocity and acceleration limits are not taken into account, this approach will tolerate violations of the position limits in general and yet provide an acceleration interval causing the position to converge to a value within the limits.

\subsection{Rendering of Serial Kinematic Chains}
\label{sec:control:rendering}

In the absence of constraints, the DT's end effector shall behave according to the $n$-DOF dynamic model
\begin{equation}
  \label{eq:control:rendering:dynamics}
  \mat{M}{\adm} \ddq{\adm}
  +
  \vek{c}{\adm}(\q{\adm}, \dq{\adm})
  =
  \mat{J}{\adm\tranT}
  (
  \vek{w}{\fhrbase}{\fdtee}
  -
  \vek{w}{\fhrbase}{\fdtee,\text{ref}}
  )
  -
  \vek{\tau}{\adm}{\text{dis}}
  +
  \vek{\tau}{\adm}{\text{dri}}
  \mathspace,
\end{equation}
which can be interpreted as non-linear admittance.
In this expression, $\vek{\tau}{\adm}{\text{dis}}$ and $\vek{\tau}{\adm}{\text{dri}}$ represent dissipative and driving torques, respectively.
The model is driven by the difference between user wrench $\vek{w}{\fhrbase}{\fdtee}$ and reference wrench $\vek{w}{\fhrbase}{\fdtee,\text{ref}}$.
In practice, rendering this dynamics equation is non-trivial, as joint limits, Cartesian limits, and singularities must be respected simultaneously.
To solve this, we leverage an HQP problem with the following tasks.
\subsubsection{Kinematic Coupling}
\label{sec:control:rendering:task:coupling}
The movement of the manipulator's end effector must coincide with that of the DT's end effector. Therefore,
\begin{equation}
  \begin{pmatrix}
    \vek{\ddot{x}}{\fhrbase}{\fhrbase\fdtee} - \vek{\ddot{x}}{\fhrbase}{\fhrbase\fhree} \\
    \vek{\ddot{\omega}}{\fhrbase}{\fhrbase\fdtee} - \vek{\ddot{\omega}}{\fhrbase}{\fhrbase\fhree}
  \end{pmatrix}
  = \vek{s}
  \label{eq:control:rendering:task:coupling}
\end{equation}
is set as an equality task with the synchronization term
\begin{equation}
  \vek{s}
  \!=\!
  \begin{pmatrix}
    k_{\text{P,tran}} (\vek{x}{\fhrbase}{\fhrbase\fdtee} - \vek{x}{\fhrbase}{\fhrbase\fhree}) \\
    k_{\text{P,rot}} \frac{1}{2} (\mat{C}{\fhrbase}{\fdtee}+\mat{C}{\fhrbase}{\fhree}) \vek{r}{\fhree}{\fhree\fdtee} \gamma^{\fhree}_{\fdtee}
  \end{pmatrix}
  \!+\!
  \begin{pmatrix}
    k_{\text{D,tran}} (\vek{\dot{x}}{\fhrbase}{\fhrbase\fdtee} - \vek{\dot{x}}{\fhrbase}{\fhrbase\fhree}) \\
    k_{\text{D,rot}} (\vek{\dot{\omega}}{\fhrbase}{\fhrbase\fdtee} - \vek{\dot{\omega}}{\fhrbase}{\fhrbase\fhree})
  \end{pmatrix}
\end{equation}
mimicking a PD pose controller.
The rotational difference term is based on the rotation axis $\vek{r}{\fhree}{\fhree\fdtee}$ and angle $\gamma^{\fhree}_{\fdtee}$ of $\mat{C}{\fhree}{\fdtee}$.
In an ideal world, $\vek{s} = \vek{0}$ holds.
However, synchronization is necessary in practice because there are errors from numerical integration and linearization which would cause the two end effectors to drift apart.
Using joint space variables, the left side of \eqref{eq:control:rendering:task:coupling} can be made compatible with \eqref{eq:control:hqp:task}, yielding
\begin{equation}
  \begin{pmatrix}
    \mat{J}{\ppu} & \mat{J}{\manip}
  \end{pmatrix}
  \begin{pmatrix}
    \ddq{\ppu} \\
    \ddq{\manip}
  \end{pmatrix}
  +
  \begin{pmatrix}
    \mat{\dot{J}}{\ppu} & \mat{\dot{J}}{\manip}
  \end{pmatrix}
  \begin{pmatrix}
    \dq{\ppu} \\
    \dq{\manip}
  \end{pmatrix}
  -
  \mat{J}{\adm} \ddq{\adm}
  -
  \mat{\dot{J}}{\adm} \dq{\adm}
  =
  \vek{s}
  \mathspace.
\end{equation}

\subsubsection{Manipulator Joint Position and Velocity Limits}
\label{sec:control:rendering:task:manipulator_limits_1}

For safety, the manipulator must not exceed its joint position and velocity limits.
Note, that acceleration limits are not considered here, as they would be overconstraining the manipulator movement at this point.
Thus, the first type of bounds from \refsec{sec:control:limits} is used in the inequality task
\begin{equation}
  \vek{l}{\manip}{\text{lb},A}(\q{\manip}, \dq{\manip}) \le \ddq{\manip} \le \vek{l}{\manip}{\text{ub},A}(\q{\manip}, \dq{\manip}) \mathspace.
\end{equation}

\subsubsection{Singularity Avoidance}
\label{sec:control:rendering:task:singularity}
Even though the manipulator was designed to have as few singularities as possible, one singularity still occurs when $q^\sing = \q{m}{[4]}+ \q{m}{[5]}$ is $0$ or $\pi$, because the axes of joint 1, 2, 3, and 6 align in this case.
To avoid this, \mbox{$0 < s_\text{min} \le q^\sing \le s_\text{max} < \pi$} is introduced.
Here, the parameters $s_\text{min}$ and $s_\text{max}$ act as a position constraint of a virtual sum joint.
This can be implemented as a task using the last constraint type from \refsec{sec:control:limits}, resulting in
\begin{equation}
  \label{eq:control:rendering:task:singularity}
  \vek{l}{\sing}{\text{lb},C}(q^\sing, \dot{q}^\sing)
  \le
  \ddot{q}^\sing
  \le
  \vek{l}{\sing}{\text{ub},C}(q^\sing, \dot{q}^\sing)
  \mathspace.
\end{equation}

\subsubsection{Cartesian Limits}
\label{sec:control:rendering:task:cartesian_limits}
Collisions of the manipulator with the supporting frame must be avoided.
Reducing the PPU's range of motion is not a reasonable option, as it would result in a small dexterous workspace due to the enormous maximal reach of the manipulator.
Therefore, the end effector position $\vek{x}{\fhrbase}{\fhrbase\fhree}$ must be limited to
$\vek{x}{\fhrbase}{\text{lb},\fhrbase\fhree} \le \vek{x}{\fhrbase}{\fhrbase\fhree} \le \vek{x}{\fhrbase}{\text{ub},\fhrbase\fhree}$.
To realize this, the Cartesian position is interpreted as a kinematic chain with three orthogonal DOF. With the same reasoning as before, the limits
\begin{equation}
  \label{eq:control:rendering:task:cartesian_limits}
  \vek{l}{\text{ee}}{\text{lb},C}(\vek{x}{\fhrbase}{\fhrbase\fhree}, \vek{\dot{x}}{\fhrbase}{\fhrbase\fhree})
  \le
  \vek{\ddot{x}}{\fhrbase}{\fhrbase\fhree}
  \le
  \vek{l}{\text{ee}}{\text{ub},C}(\vek{x}{\fhrbase}{\fhrbase\fhree}, \vek{\dot{x}}{\fhrbase}{\fhrbase\fhree})
\end{equation}
have to be maintained.
To make this compatible with the decision variable, the differential kinematics
\begin{equation}
  \vek{\ddot{x}}{\fhrbase}{\fhrbase\fhree}
  \!=\!
  \begin{pmatrix}
    \mat{J}{\ppu}{[1:3,:]}\! & \!\mat{J}{\manip}{[1:3,:]}
  \end{pmatrix}
  \!\!
  \begin{pmatrix}
    \ddq{\ppu} \\
    \ddq{\manip}
  \end{pmatrix}
  +
  \begin{pmatrix}
    \mat{\dot{J}}{\ppu}{[1:3,:]}\! & \!\mat{\dot{J}}{\manip}{[1:3,:]}
  \end{pmatrix}
  \!\!
  \begin{pmatrix}
    \dq{\ppu} \\
    \dq{\manip}
  \end{pmatrix}
\end{equation}
can be used to substitute $\vek{\ddot{x}}{\fhrbase}{\fhrbase\fhree}$ in \eqref{eq:control:rendering:task:cartesian_limits}.
Likewise, the position of the elbow joint, which coincides with manipulator joint 4, is constrained to the same Cartesian range of motion with a separate task.
Further Cartesian limits are unnecessary, as safety margins are in place and the collision of joint 2 with the frame is mechanically impossible.

\subsubsection{Admittance Position and Velocity Limits}
\label{sec:control:rendering:task:admittance_limits_1}

Analog to \refsec{sec:control:rendering:task:manipulator_limits_1}, the joint position and velocity limits of the admittance
are added using the task
\begin{equation}
  \label{eq:control:rendering:task:admittance_limits_1}
  \vek{l}{\adm}{\text{lb},A}\!(\q{\adm}, \dq{\adm}) \le \ddq{\adm} \le \vek{l}{\adm}{\text{ub},A}\!(\q{\adm}, \dq{\adm})
  \mathspace.
\end{equation}
The priority of this task is lower than that of the manipulator limits, as violating admittance limits will impair the haptic quality but not risk damaging HapticGiant.

\subsubsection{Admittance and Manipulator Acceleration Limits}
\label{sec:control:rendering:task:acceleration_limits}

In case of a limited maximum acceleration for the admittance or the manipulator joints, the inequality constraints
\begin{eqnarray}
  \label{eq:control:rendering:task:manipulator_limits_2}
  \vek{l}{\manip}{\text{lb},B}(\q{\manip}, \dq{\manip}) \le& \ddq{\manip} &\le \vek{l}{\manip}{\text{ub},B}(\q{\manip}, \dq{\manip}) \mathspace\mathspace\mathspace \text{and} \\
  \label{eq:control:rendering:task:admittance_limits_2}
  \vek{l}{\adm}{\text{lb},B}(\q{\adm}, \dq{\adm}) \le& \ddq{\adm} &\le \vek{l}{\adm}{\text{ub},B}(\q{\adm}, \dq{\adm})
\end{eqnarray}
can be added in this order as separate tasks.
Note, that especially the admittance acceleration limit will be violated occasionally, e.g., when the manipulator hits a limit of higher priority.
This is the reason why the position and velocity limits are split into separate tasks for both, the manipulator and the admittance.

\subsubsection{Addmitance Dynamic Behavior}
\label{sec:control:rendering:task:dynamics}
So far, the formulation neither responds to user-exerted forces nor behaves admittance-like.
To add this, the dynamic behavior \eqref{eq:control:rendering:dynamics} is added as equality task
\begin{equation}
  \label{eq:control:rendering:task:dynamics}
  \ddq{\adm}
  \!-\!
  ({\mat{M}{\adm}})^{\!-1}
  \vek{\tau}{}{\text{con}}
  \!\!=\!
  ({\mat{M}{\adm}})^{\!-1}
  \!\!
  \left(
  \mat{J}{\adm\tranT}
  \!
  (
  \vek{w}{\fhrbase}{\fdtee}
  \!\!-\!
  \vek{w}{\fhrbase}{\fdtee,\text{ref}}
  )
  \!\!-\!
  \vek{\tau}{\adm}{\text{dis}}
  \!\!+\!
  \vek{\tau}{\adm}{\text{dri}}
  \!\!-\!
  \vek{c}{\adm}
  \right)
\end{equation}
with the addition of the total constraint torque $\vek{\tau}{}{\text{con}}$.
The equation is purposefully formulated in joint acceleration space, as this resulted in better behavior for poorly conditioned $\mat{M}{\adm}$.
All constraints are inactive when none of the previous inequalities hit their lower or upper limit.
In this case, the slack variables are zero and the rendering already does what is desired, i.e., it behaves \emph{dynamically consistent}. 
However, when an inequality constraint is hit while $\vek{\tau}{}{\text{con}} = \vek{0}$, the HQP solver will find a joint acceleration that minimizes the norm of the acceleration error, which differs from the physically expected behavior.
To fix this, the generalized constraint torque must be applied in a way that behaves like a physical barrier.
For example, assume that the last manipulator joint hits its upper position limit.
If a real mechanical limit was hit, a constraint torque would occur at the last joint, inducing a torque on the mechanically coupled admittance joints and eventually causing dynamically consistent behavior.
To reproduce this algorithmically, the constraint torque is set to a sum, where each term corresponds to one of the previously introduced constraints with their respective constraint forces/torques $\vek{\alpha}{}{\text{con}}$.

\begin{enumerate}[label=(\alph*)] 
  \item The first term is simply $\vek{\tau}{\adm}{\text{con}} = \vek{\alpha}{\adm}{\text{con}} \in \mathbb{R}^n$ as hitting an admittance joint limit will cause a counter-torque acting individually on each admittance joint.
  \item Likewise, manipulator limits will create the torque \mbox{$\vek{\alpha}{\manip}{\text{con}} \in \mathbb{R}^6$} acting individually on each joint.
        This torque is then transformed into the admittance torque space using
        \begin{equation}
          \label{eq:control:rendering:task:dynamics:constraint_torque_manip}
          \vek{\tau}{\manip}{\text{con}} =
          {\mat{J}{\adm}}\tran
          \left({\mat{J}{\manip}}\tran\right)^{-1}
          \vek{\alpha}{\manip}{\text{con}}
          \mathspace.
        \end{equation}
  \item When the manipulator's singularity is hit, a constraint torque $\alpha^{\sing}_{\text{con}} \in \mathbb{R}$ pushing the manipulator away from the singularity is necessary. This is achieved by adding the term
        \begin{equation}
          \vek{\tau}{\sing}{\text{con}}
          =
          (\mat{J}{\adm}{[4:6,:]})\tran
          \vek{r}^{\fhrbase}_\text{elbow}
          \alpha^{\sing}_{\text{con}}
          \mathspace,
        \end{equation}
        where $\vek{r}^{\fhrbase}_\text{elbow}$ is the instantaneous rotation axis of joints~4 and~5.
  \item The Cartesian end effector constraints are reflected by the Cartesian constraint force $\vek{\alpha}{\ee}{\text{con}} \in \mathbb{R}^3$, which is transformed into the admittance torque space using
        \begin{equation}
          \vek{\tau}{\ee}{\text{con}} =
          (\mat{J}{\adm}{[1:3,:]})\tran
          \vek{\alpha}{\ee}{\text{con}}
          \mathspace.
        \end{equation}
  \item The Cartesian elbow constraints are formulated similarly, but with two transformations. First, the Cartesian constraint forces $\vek{\alpha}{\elbow}{\text{con}} \in \mathbb{R}^3$ are converted into manipulator torques using the elbow Jacobian $\mat{J}{\elbow}$. Then, these are transformed to admittance joint torques as in \eqref{eq:control:rendering:task:dynamics:constraint_torque_manip}, yielding
        \begin{equation}
          \vek{\tau}{\elbow}{\text{con}} =
          {\mat{J}{\adm}}\tran
          ({\mat{J}{\manip}}\tran)^{-1}
          (\mat{J}{\text{elbow}}{[1:3,:]})\tran
          \vek{\alpha}{\elbow}{\text{con}}
          \mathspace.
        \end{equation}
        In summary, the decision variables of the HQP formulation are
        \begin{equation}
          \setlength\arraycolsep{3pt}
          \vek{x}{\tranT} =
          \begin{pmatrix}
            \ddq{\manip\tranT}                     &
            \ddq{\adm\tranT}                       &
            \vek{\alpha}{\adm\tranT}{\text{con}}   &
            \vek{\alpha}{\manip\tranT}{\text{con}} &
            \alpha^{\sing}_{\text{con}}            &
            \vek{\alpha}{\ee\tranT}{\text{con}}    &
            \vek{\alpha}{\elbow\tranT}{\text{con}}
          \end{pmatrix}
          .
        \end{equation}
\end{enumerate}

\subsubsection{Constraint Torque Range}
\label{sec:control:rendering:task:constraint_torque_range}

The goal of dynamically consistent behavior is not yet achieved in all situations because the different constraint torque contributions can take arbitrary values.
To avoid this, each individual entry $\alpha$ of the different $\vek{\alpha}{}{\text{con}}$ is bound according to
\begin{equation}
  \label{eq:control:rendering:task:constraint_torque_range:cases}
  \begin{cases}
    -\infty \le \alpha \le 0, & \text{if upper, but no lower limit active}, \\
    0 \le \alpha \le \infty,  & \text{if lower, but no upper limit active}, \\
    0 \le \alpha \le 0,       & \text{if no limit active}.
  \end{cases}
\end{equation}
The last case is intentionally not formulated as $\alpha = 0$, since this would require switching between equality and inequality constraints at runtime.
To decide which limits are active, the optimization result from the previous iteration is reinserted into the inequality constraints from the tasks in \refsec{sec:control:rendering:task:manipulator_limits_1} to \ref{sec:control:rendering:task:acceleration_limits}.
Similar to an active-set strategy, a limit is considered active if the equality holds for a given \enquote{less or equal than} constraint.
This means that dynamic consistency is not given whenever a limit switch occurs. In practice, however, the control frequency is high enough for this effect to be unnoticeable.

\subsubsection{Constraint Torque Norm}
\label{sec:control:rendering:task:constraint_torque_norm}

In the last task, all generalized constraint torques $\vek{\alpha}{}{\text{con}}$ are set to zero to minimize the overall used constraint torques in cases where ambiguities still exist.
This cannot be added to the previous task, as the optimizer would find a trade-off between equality and inequality slack instead of prioritizing the constraint torque ranges.
Furthermore, this step formally ensures that the final solution space of the HQP problem has dimension zero.
\reftab{tab:control:rendering:task_summary} summarizes all tasks for the proposed hierarchical force control scheme.

\begin{table}
  \centering
  \renewcommand*{\arraystretch}{1.2}
  \begin{tabular}{ccl}
    \toprule
    Priority & Type & Task
    \\ \midrule
    1        & EQ   & Kinematic coupling                           \\
    2        & IQ   & Manipulator joint angle \& velocity limits   \\
    3        & IQ   & Singularity avoidance                        \\
    4        & IQ   & Cartesian end effector limits                \\
    5        & IQ   & Cartesian elbow limits                       \\
    6        & IQ   & Admittance joint angle \& velocity limits    \\
    7        & IQ   & Manipulator joint acceleration limits        \\
    8        & IQ   & Admittance joint acceleration limits         \\
    9        & EQ   & Admittance dynamics                          \\
    10       & IQ   & Generalized constraint torques range         \\
    11       & EQ   & Generalized constraint torques normalization \\
    \bottomrule
  \end{tabular}
  \caption{Summary of the tasks in the HQP formulation. The type states whether the task contains an equality (EQ) or an inequality (IQ) constraint.}
  \label{tab:control:rendering:task_summary}
\end{table}

\subsection{Prepositioning Unit}
\label{sec:control:ppu}

To optimize the workspace utilization and the manipulability of HapticGiant, it is necessary to move the velocity-controlled PPU alongside the manipulator, whose movement is calculated in the hierarchical optimization.
By interpreting the actual PPU movement $\dq{\ppu}$ and $\ddq{\ppu}$ as a known disturbance, the PPU position can be optimized independently without altering the above HQP formulation.

With the desired PPU position $\q{\ppu}{\text{ref}}$ and the actual PPU position $\q{\ppu}$, the commanded PPU velocity
\begin{equation}
  \dq{\ppu}{\text{cmd}} =
  k_\text{PPU} (\q{\ppu}{\text{ref}} - \q{\ppu})
\end{equation}
can be formulated as a P-controller.
Even with physical limit switches, it is crucial to avoid reaching the PPU limits at high speeds to ensure a smooth user experience.
This is achieved by applying the velocity bounds
\begin{eqnarray}
  \dq{\ppu}{\text{lb},\text{ref}} &=& \nicefrac{1}{\Delta t}(\q{\ppu}{\text{lb}}- \q{\ppu}) \quad\text{and}\\
  \dq{\ppu}{\text{ub},\text{ref}} &=& \nicefrac{1}{\Delta t}(\q{\ppu}{\text{ub}}- \q{\ppu}) \mathspace,
\end{eqnarray}
derived from a constant velocity model, to $\dq{\ppu}{\text{cmd}}$ to finalize the setpoint.

Without a specific application in mind, we suggest two goals for the PPU position:
\begin{enumerate}
  \item The manipulability of the SCARA manipulator composed of joints 1 and 2 shall be maximized, i.e., $\q{\manip}{[2]} = \pi/2$.
  \item The ground projection of the link between joints 4 and 5 shall be aligned with the line connecting joints 1 and 3. This yields $\q{\manip}{[3]} = -\arctan(\SI{0.53}{\meter} / \SI{0.67}{\meter})$.
\end{enumerate}
The sum of the first three joint angles has to be constant for a given end effector orientation. Hence, $\q{\manip}{[1]}$ is selected accordingly.
To obtain the optimal PPU position, the end effector position $\vek{x}{\fhrbase}{\fhrbase\fhree}$ is calculated first with the current manipulator configuration.
Then, the auxiliary end effector position $\vek{x}{\fhrbase}{\fhrbase\fhree,\text{aux}}$ is determined with the updated values for $\q{\manip}{[1]}$ to $\q{\manip}{[3]}$.
The desired PPU position is then
\begin{equation}
  \q{\ppu}{\text{ref}}
  =
  \q{\ppu}
  +
  \vek{x}{\fhrbase}{\fhrbase\fhree[1:2]}
  -
  \vek{x}{\fhrbase}{\fhrbase\fhree,\text{aux}[1:2]}
  \mathspace.
\end{equation}

\subsection{Low-Level Control}
\label{sec:control:lowlevel_control}

The presented optimization scheme provides joint accelerations for the admittance and the manipulator, which are both numerically integrated using a constant acceleration system model.
The resulting joint state of the admittance is used to visualize the DT.
The joint state of the manipulator is handed over to a tracking controller, which has feedforward terms for gravitational load, the estimated user wrench, and a friction model incorporating viscous, Coulomb, as well as load-dependent friction terms.
The commanded prepositioning velocity is simply forwarded to the inverters and their internal control loops.

\begin{figure*}
  \newcommand{\Je}{$\mat{J}{\manip\tranT}{}$}
  \newcommand{\taucmd}{$\vek{\tau}{\manip}{\text{cmd}}$}
  \newcommand{\vppucmd}{$\vek{\dot{q}}{\ppu}{\text{cmd}}$}
  \newcommand{\wmeas}{$\vek{w}{\fhrbase}{\fdtee}$}
  \newcommand{\wcmd}{$\vek{w}{\fhrbase}{\fdtee,\text{ref}}$}
  \newcommand{\ddqppu}{$\vek{\ddot{q}}{\ppu}$}
  \newcommand{\dqppu}{$\vek{\dot{q}}{\ppu}$}
  \newcommand{\qppu}{$\vek{q}{\ppu}$}
  \newcommand{\ddqa}{$\vek{\ddot{q}}{\adm}$}
  \newcommand{\dqa}{$\vek{\dot{q}}{\adm}$}
  \newcommand{\qa}{$\vek{q}{\adm}$}
  \newcommand{\ddqm}{$\vek{\ddot{q}}{\manip}$}
  \newcommand{\dqm}{$\vek{\dot{q}}{\manip}$}
  \newcommand{\qm}{$\vek{q}{\manip}$}

  \centering
  \def\svgwidth{\textwidth}
  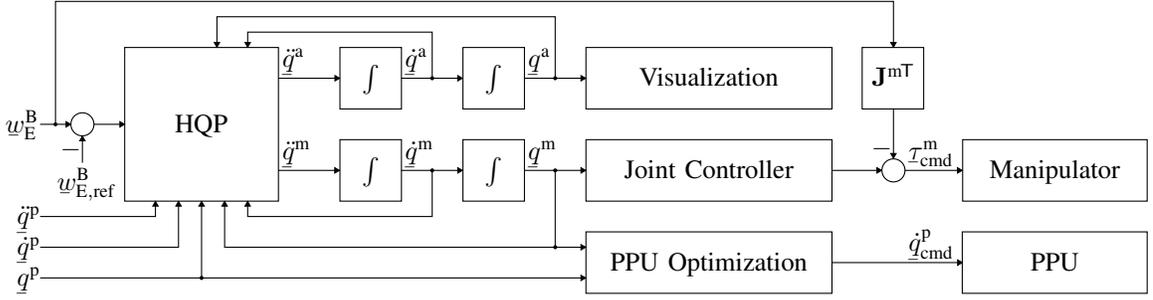


  \caption{Block diagram of the control architecture. The quantity $\vek{q}{\ppu}$ and its derivatives originate from the state estimation.}
  \label{fig:control:overview}
\end{figure*}

To keep this paper concise, we deliberately exclude further details about the low-level controllers.
This decision is backed by the fact that any tracking controller with suitable inputs and outputs can be used for tracking the PPU and the manipulator state. The overall control scheme is summarized in \reffig{fig:control:overview}.

\section{Experiments}
\label{sec:experiments}

In the following, three experiments are performed with HapticGiant to prove the effectiveness and demonstrate potential applications of the overall system.
For a better understanding, we also provide a supplementary video.%
\footnote{{\tt \url{https://youtu.be/q9BG-ZEPCs0}}}
The presented control scheme was implemented in C++ as a module in the software architecture from \refsec{sec:architecture} using \emph{pinocchio}~\cite{carpentierPinocchioLibraryFast2019c} for kinematic and dynamic calculations and \emph{QuadProg++}~\cite{LiuqQuadProgppLibrary} as base for a custom HQP solver.
The control frequency has been set to \SI{1}{\kilo\hertz} on a system with an \emph{Intel Core i5-8600} CPU.

The scope of this paper is limited to the overall system and high-level control aspects.
For this reason, we do not provide a quantitative end-to-end evaluation of the haptic performance, since the low-level controllers, which strongly affect the overall system performance, are subject of ongoing research.
As known from literature~\cite{newmanStabilityPerformanceLimits1992a}, there is always a passive environment (e.g., stiff spring), where contact instability occurs under linear admittance control as soon as the rendered admittance is smaller than the passive device inertia. 
This leads to the fact that the stability limits of the overall system are strongly dependent on the tracking controllers and there is no stable linear controller under all conditions.
The presented admittance control scheme is no exception to this rule. 
Therefore, we intentionally omit stability considerations.
Several methods have been proposed in the literature to address this issue, either through passivity-based controllers~\cite{cordoniVariableStochasticAdmittance2020,schindlbeckUnifiedPassivitybasedCartesian2015} or by adjusting control parameters when oscillations arise~\cite{landiAdmittanceControlParameter2017,landiVariableAdmittanceControl2017}.
In the case of HapticGiant, the manipulator's elasticity---stemming from the series elastic actuators and the finite link stiffness---causes noticeable mechanical oscillations during manual excitation when the joint brakes are applied.
This introduces additional complex poles that complicate the tracking controller design, and a full treatment of these concerns lies beyond the scope of this paper.

\subsection{Cartesian Admittance}

\begin{figure*}
  \centering
  \newlength{\stripimgwidth}
  \setlength{\stripimgwidth}{0.1415\linewidth}
  \subfigure[User walking through the workspace. The frames were extracted \SI{1}{s} apart.
  ]{
    \includegraphics[width=\stripimgwidth]{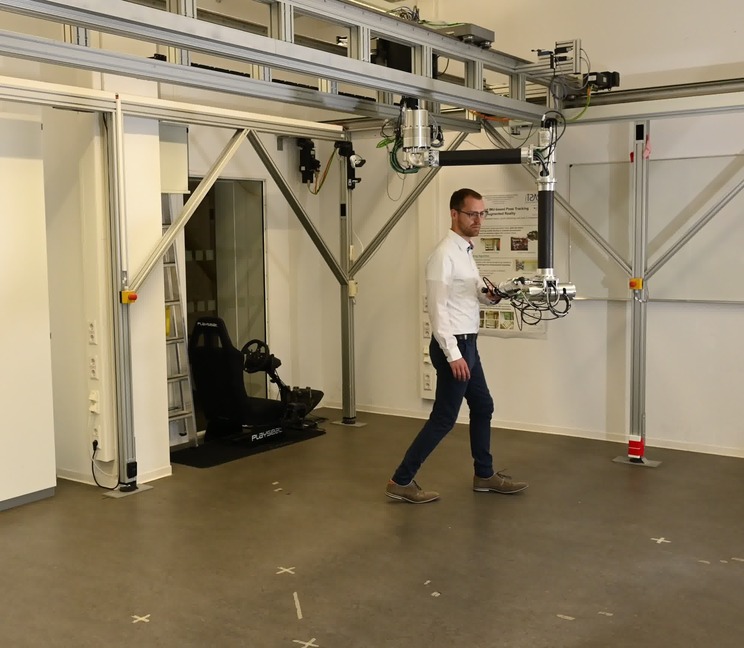}%
    \includegraphics[width=\stripimgwidth]{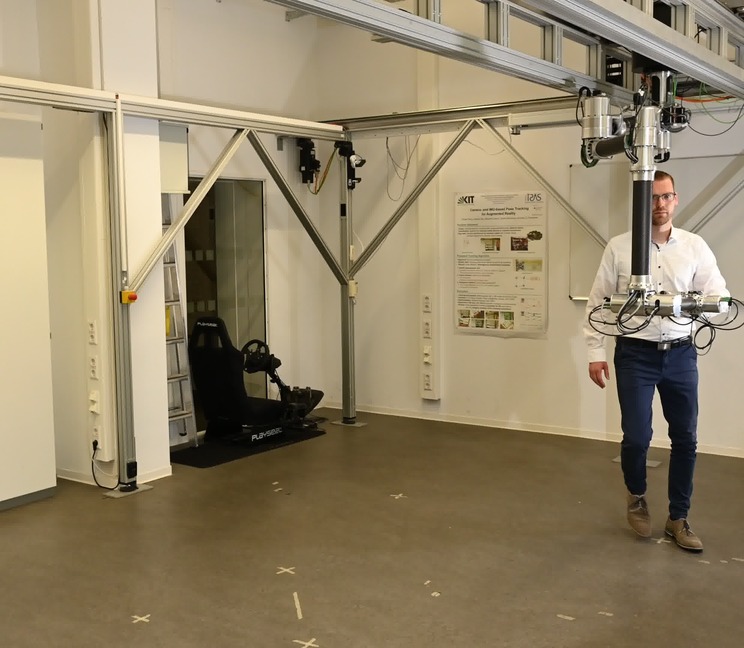}%
    \includegraphics[width=\stripimgwidth]{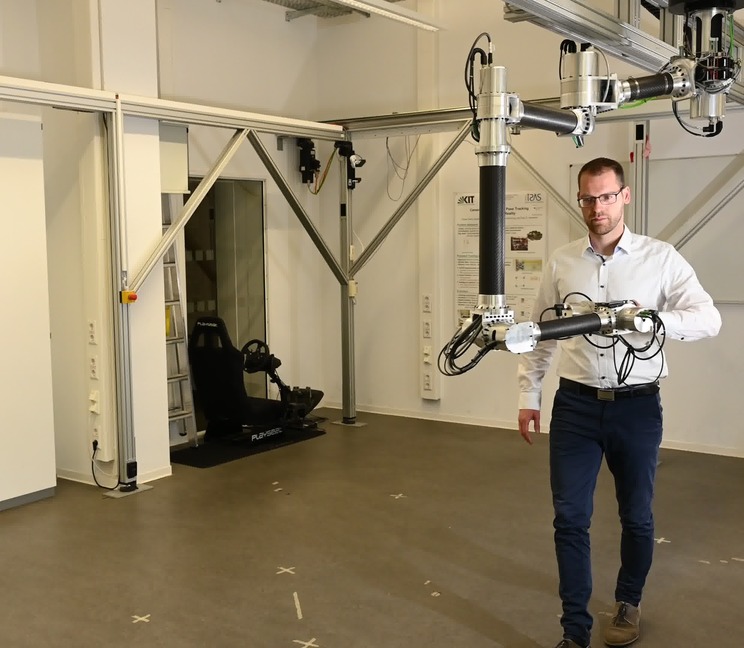}%
    \includegraphics[width=\stripimgwidth]{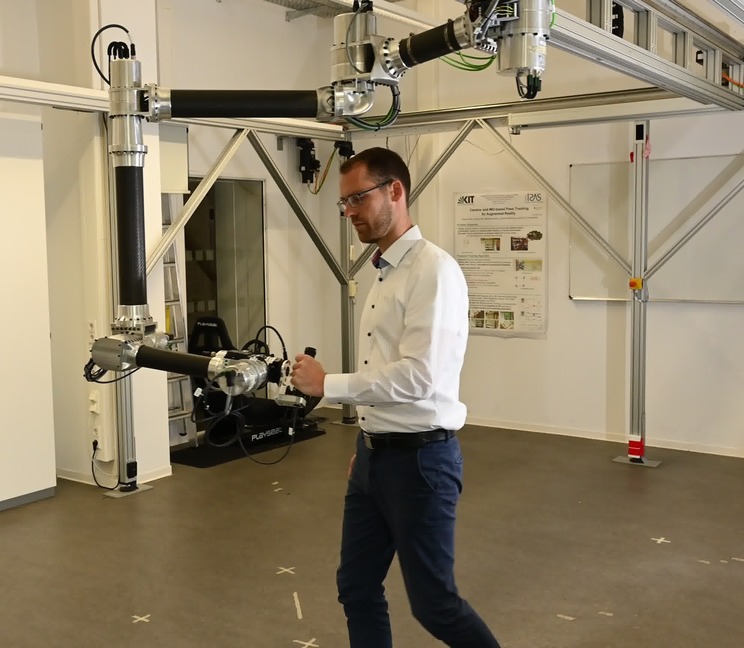}%
    \includegraphics[width=\stripimgwidth]{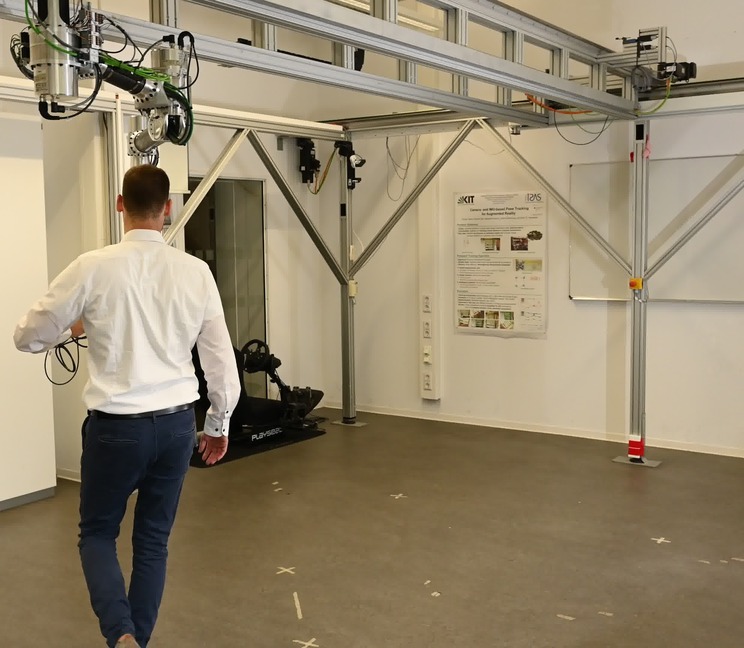}%
    \includegraphics[width=\stripimgwidth]{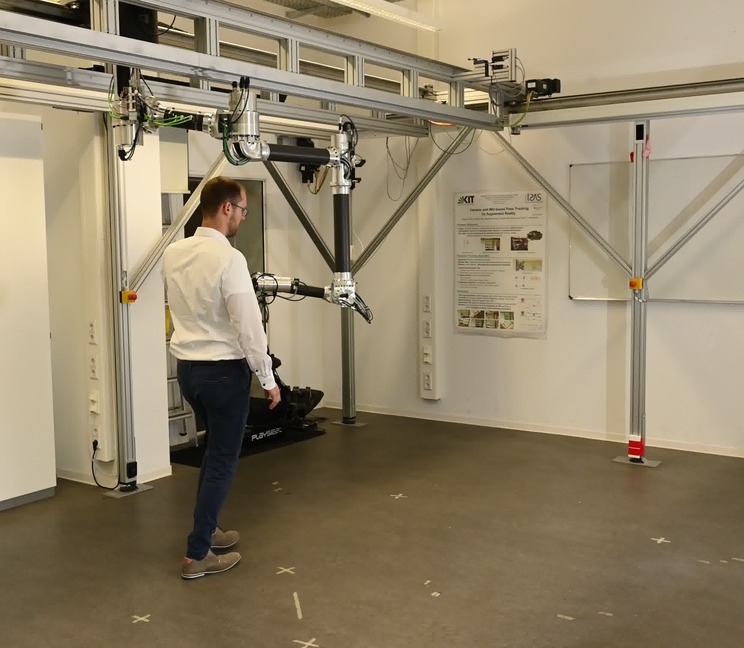}%
    \includegraphics[width=\stripimgwidth]{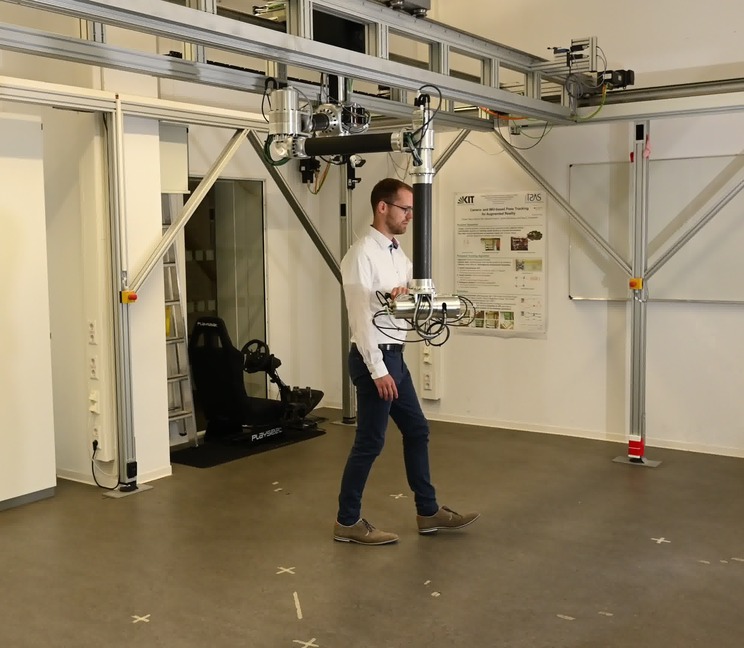}%
    \label{fig:experiments:admittance:demo:free_walking}
  }

  \subfigure[A selection of reachable hand and arm poses demonstrating the dexterity of HapticGiant's workspace.
  ]{
    \includegraphics[width=\stripimgwidth]{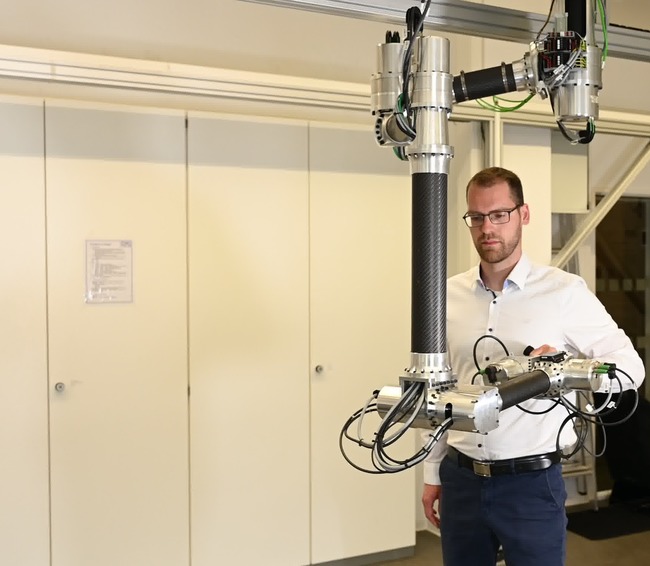}%
    \includegraphics[width=\stripimgwidth]{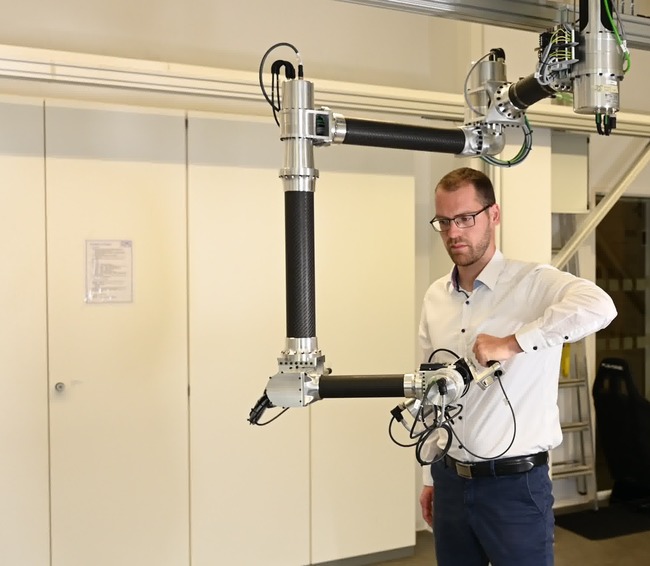}%
    \includegraphics[width=\stripimgwidth]{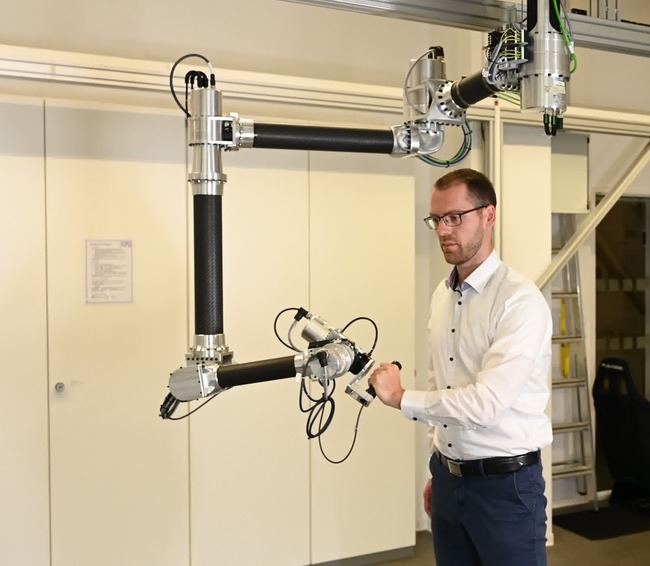}%
    \includegraphics[width=\stripimgwidth]{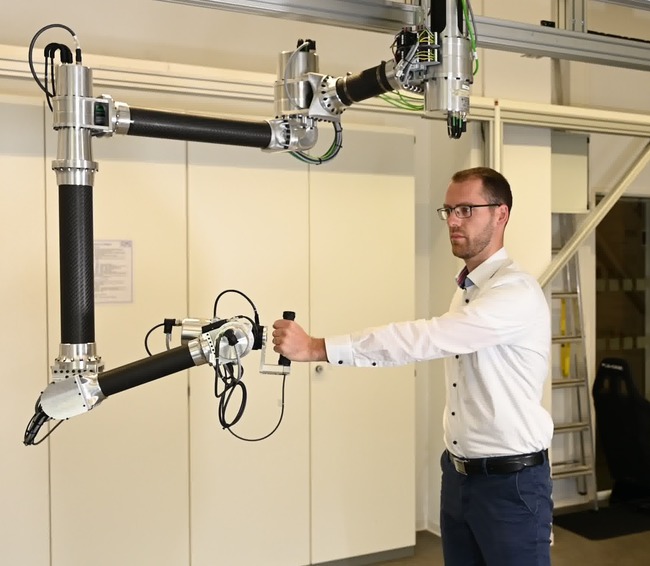}%
    \includegraphics[width=\stripimgwidth]{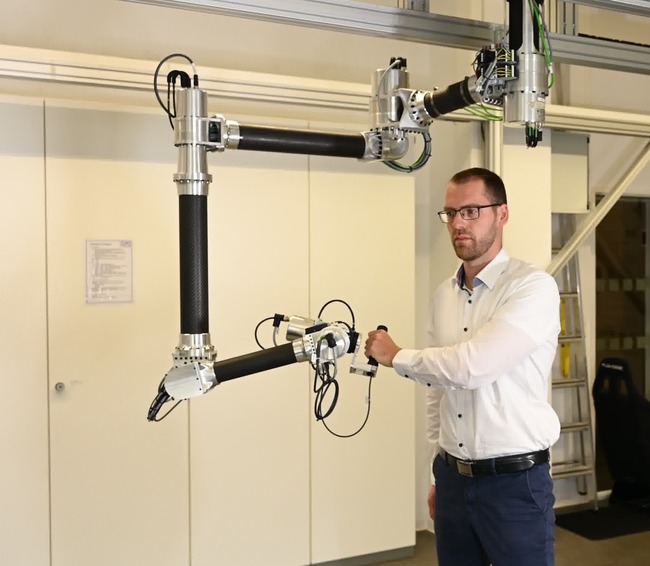}%
    \includegraphics[width=\stripimgwidth]{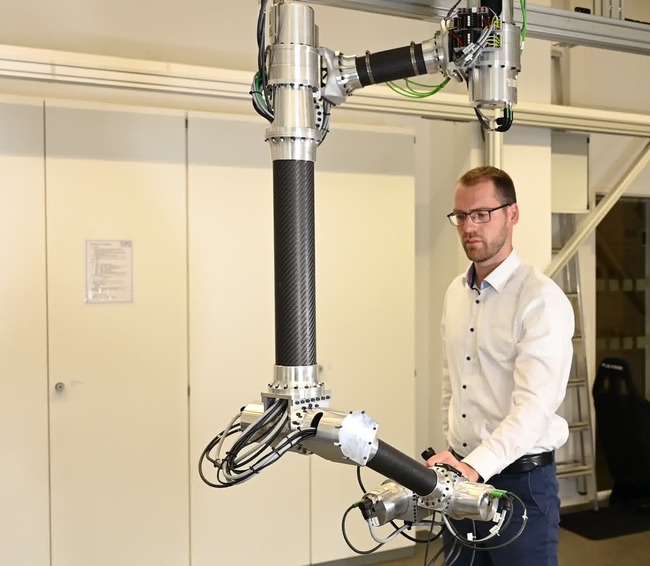}%
    \includegraphics[width=\stripimgwidth]{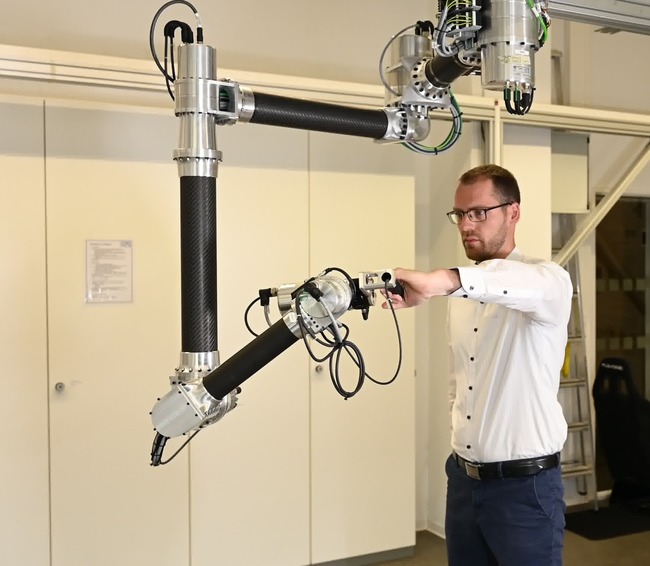}%
    \label{fig:experiments:admittance:demo:arm_poses}
  }
  \caption{User interactions during rendering of the 6 DOF Cartesian admittance.}
  \label{fig:experiments:admittance:demo}
\end{figure*}

\begin{figure}
\newcommand{\scaleAttmitance}{1.}
  \centering
  \includegraphics[width=\scaleAttmitance\linewidth,trim={0 0 0 37mm},clip]{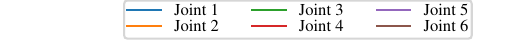}
  \subfigure[
    Manipulator joint reference angles.
    The lower limits of joints 2 and 4 as well as the upper limits of joints 5 and 6 are coincident.
  ]{
    \includegraphics[width=\scaleAttmitance\linewidth,trim={0 7.5mm 0 0},clip]{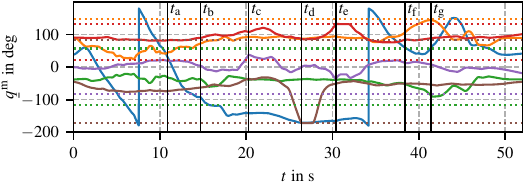}
    \label{fig:experiments:admittance:q_manip}
  }
  \subfigure[Singularity measure and its limits.]{
    \includegraphics[width=\scaleAttmitance\linewidth,trim={0 7.5mm 0 0},clip]{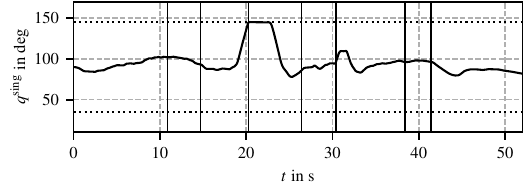}
    \label{fig:experiments:admittance:q_sing}
  }
  \subfigure[Norm of the user applied force and torque.]{
    \includegraphics[width=\scaleAttmitance\linewidth]{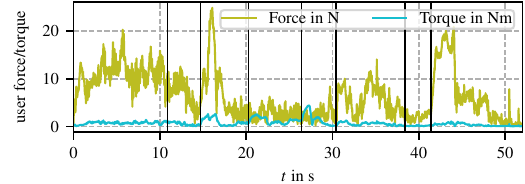}
    \label{fig:experiments:admittance:user_wrench}
  }
  \subfigure[
    Top view of the Cartesian motion of the end effector, elbow, and PPU. The limits for the elbow and the end effector are coincident.
    For comprehensibility, only the data between $t = \SI{10}{\second}$ and $t = \SI{45}{\second}$ is plotted.
  ]{
    \includegraphics[width=\scaleAttmitance\linewidth]{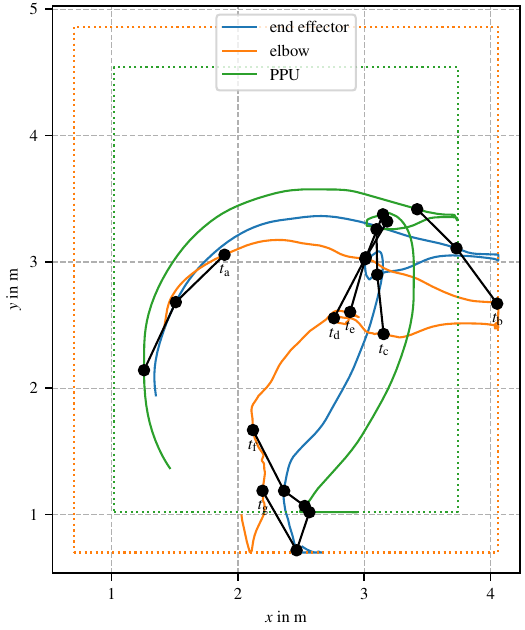}
    \label{fig:experiments:admittance:cart_2d}
  }

  \caption{Demonstration of the HQP formulation for a 6 DOF Cartesian admittance. Times of interest are highlighted with black lines, while limits are marked with dotted lines.}
  \label{fig:experiments:admittance}
\end{figure}

The DT admittance formulation can be used to render a Cartesian free space admittance with a total of 6 DOF, constrained only by the mechanical limitations of HapticGiant.
This means that the user can walk around freely, while a variety of hand and arm poses are covered as demonstrated in \reffig{fig:experiments:admittance:demo}.

In the following, the exemplary run from \reffig{fig:experiments:admittance} is explained using the highlighted points of interest.
In the beginning, the user walks in a circular motion without feeling any active constraints.
In \reffig{fig:experiments:admittance:q_manip}, a full turn of joint 1 is completed at time~$t_\text{a}$, proving the effectiveness of the slip ring and its handling on the control side.
At~$t_\text{b}$ in \reffig{fig:experiments:admittance:cart_2d}, the center of joint 4, i.e., the elbow, hits its Cartesian limit.
As a result, the elbow motion in the $x$-direction of the PPU is inhibited, but the end effector continues to move.
However, a constraint is imposed on its motion, resulting in rising user forces and torques as seen in \reffig{fig:experiments:admittance:user_wrench}.
Shortly thereafter, the end effector also hits the Cartesian limit, resulting in a peak of the resistance felt by the user.
When the user hits the singularity constraint
at~$t_\text{c}$, the controller ensures that $q^{\sing}$ does not exceed the desired limits as depicted in \reffig{fig:experiments:admittance:q_sing}.
The constrained angular motion of the end effector implies an increase in the torque perceived by the user.
Another reason for a user-perceived constraint can be found at~$t_\text{d}$ in \reffig{fig:experiments:admittance:q_manip}.
Here, the user hits the lower angular limit of joint~6, which is successfully maintained by the control algorithm without affecting other parts of the manipulator.
At~$t_\text{e}$, joint~4 reaches its upper position limit, meaning that the height of the end effector cannot be increased any further.
This causes a sudden jump in the perceived user force, which effectively limits the height.
The handling of the PPU limits is demonstrated at~$t_\text{f}$ in \reffig{fig:experiments:admittance:cart_2d}.
While the PPU comes to a full stop in $y$-direction, movements in $x$-direction are still possible.
In addition, the user force and torque remain low, indicating that the user is still unconstrained.
To facilitate this, joint~2 starts to move away from its desired $\SI{90}{\degree}$ position.
At~$t_\text{g}$, joint~2 stops at its upper limit to prevent a self-collision.
As a result, the free space admittance rendering becomes constrained to a manipulator with effectively 5~DOF, which is reflected by an increasing user force.
In the example, the user then causes a rapid rotation of joint~1, until the end effector and the elbow hit the Cartesian limit.
%
In addition to the justified peaks of the user force after $t_\text{b}$ and~$t_\text{g}$, \reffig{fig:experiments:admittance:user_wrench} also shows an increased force level in the sequence before~$t_\text{b}$ as well as between $t_\text{e}$ and~$t_\text{f}$. 
Note that this is due to the inability of the PPU position P-controller~(\refsec{sec:control:ppu}) and subsequent low-level controllers to fully compensate for the dynamics of the high-inertia PPU. This can be seen in the time periods mentioned because large PPU motion is required. In the other time periods, with less PPU motion, this has no significant effect. 



In this example, the mean time between two controller iterations was $\SI{1003}{\micro\second}$, indicating that the scheduling of Preempt-RT works quite well.
The computation time of the proposed controller was \SI{529}{\micro\second} on average with a \num{99.9}th percentile of \SI{616}{\micro\second} and a maximum of \SI{707}{\micro\second}.
On average, \num{92.6} \% of the computation time was spent solving the HQP problem.

\subsection{Excavator}

\begin{figure}
  \definecolor{colorswing}{HTML}{1F77B4} 
  \definecolor{colorboom}{HTML}{ff7f0e} 
  \definecolor{colorstick}{HTML}{2ca02c} 
  \definecolor{colorbucket}{HTML}{d62728} 
  \centering
  \subfigure[Excavator with joint definitions. The order of the joints is {\color{colorswing}\emph{swing}}, {\color{colorboom}\emph{boom}}, {\color{colorstick}\emph{stick}}, and {\color{colorbucket}\emph{bucket}}.]{
    \begin{tikzpicture}
      \node[anchor=south west,inner sep=0] (image) at (0,0){
        \includegraphics[width=0.44\columnwidth]{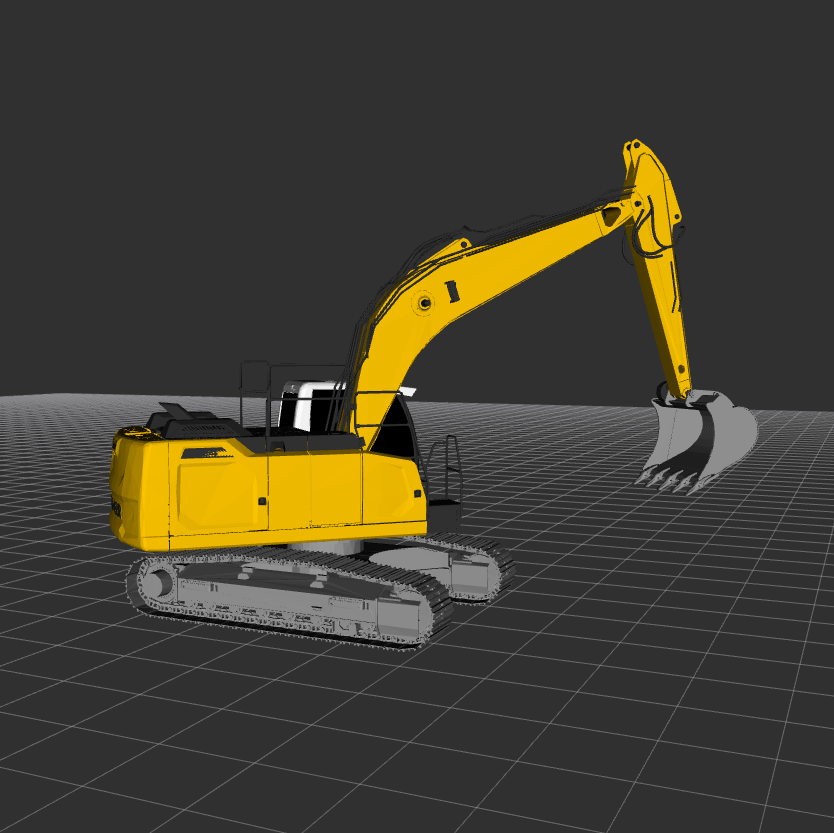}
      };
      \begin{scope}[x={(image.south east)},y={(image.north west)}, line width=1.5pt, every node/.style={fill=white, fill opacity=1, text opacity=1, inner sep=1pt}]
        \draw[colorswing, ->] (0.32,0.33) arc (-140:-30:0.1);
        \draw[colorboom, ->] (0.45,0.4) arc (280:80:0.15);
        \draw[colorstick, <-] (0.9,0.75) arc (15:135:0.15);
        \draw[colorbucket, ->] (0.95,0.55) arc (15:-100:0.15);
      \end{scope}
    \end{tikzpicture}
    \label{fig:experiments:scenarios:excavator}
  }%
  \subfigure[Door. The order of the joints is \emph{hinge} and \emph{handle}.]{
    \includegraphics[width=0.44\columnwidth]{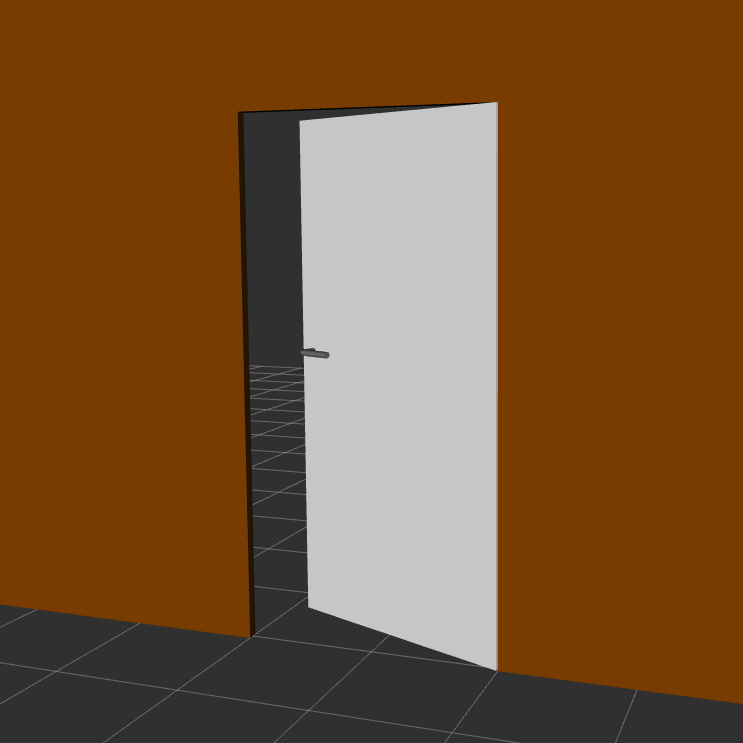}
    \label{fig:experiments:scenarios:door}
  }
  \caption{DTs used during experiments with the proposed control scheme.}
  \label{fig:experiments:scenarios}
\end{figure}
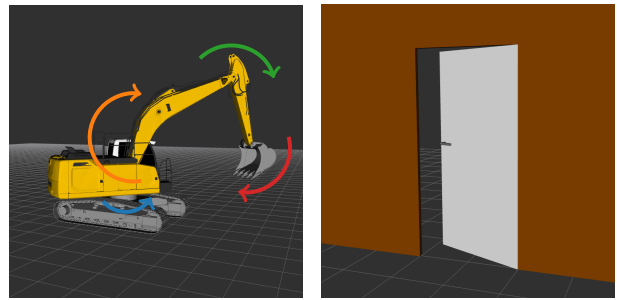

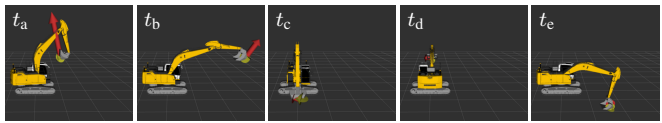
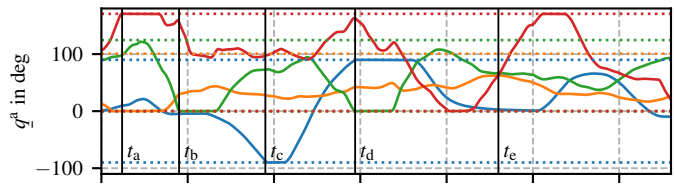
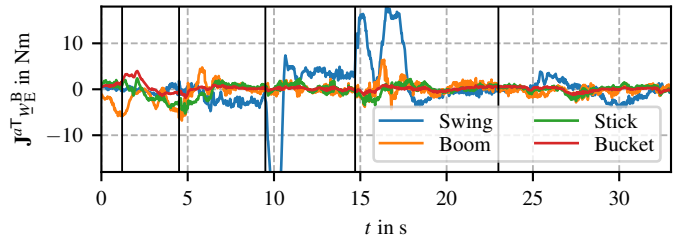
\begin{figure}
    \newcommand{\scaleExcavator}{1.}
  \centering
  \subfigure[
    Visualization of the DT at the highlighted times.
  ]{
    \newcommand{\thumbnailbox}[1]{%
      \begin{tikzpicture}
        \footnotesize%
        \node[anchor=south west,inner sep=0] (image) at (0,0) {\includegraphics[width=0.191\columnwidth]{figures/experiment_excavator/cropped_scaled/t_#1.png}};%
        \node[anchor=north west, text=white] at (image.north west) {$t_\text{#1}$};%
      \end{tikzpicture}%
    }
    \thumbnailbox{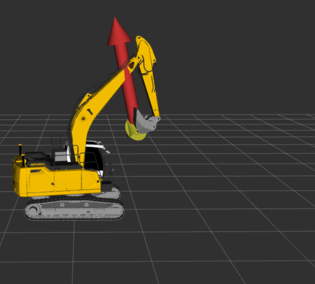}\,%
    \thumbnailbox{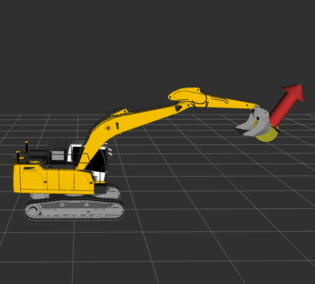}\,%
    \thumbnailbox{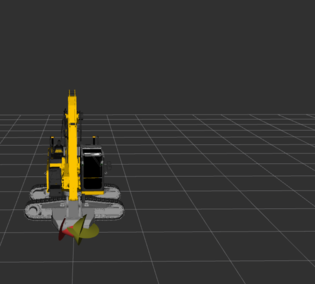}\,%
    \thumbnailbox{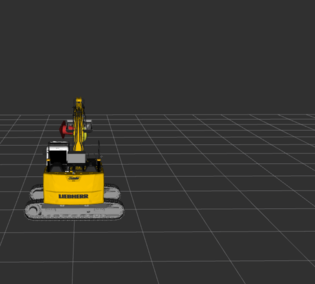}\,%
    \thumbnailbox{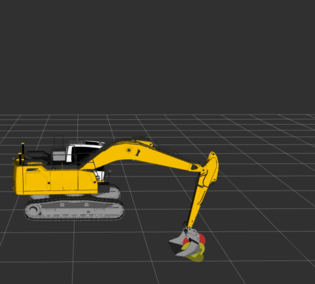}\,%
    \label{fig:experiments:excavator:miniatures}
  }

  \subfigure[
    DT joint angles with limits plotted as dotted lines.
    The lower limits of the boom, stick, and bucket joint are coincident.
  ]{
    \includegraphics[width=\scaleExcavator\linewidth,trim={0 7.5mm 0 0},clip]{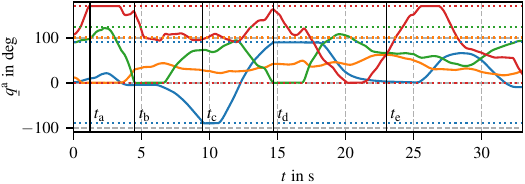}
    \label{fig:experiments:excavator:q_adm}
  }
  \subfigure[
    User wrench transformed into admittance joint space.
  ]{
    \includegraphics[width=\scaleExcavator\linewidth]{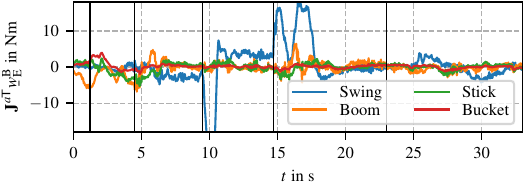}
    \label{fig:experiments:excavator:projected_user_wrench}
  }
  \caption{
    Applied user torques and resulting DT movement for the excavator application.
    Times of interest are marked with black lines.
  }
  \label{fig:experiments:excavator}
\end{figure}

In the second experiment, the user interacts with the DT of the hydraulic excavator from \reffig{fig:experiments:scenarios:excavator} by moving its bucket, which is coupled to the manipulator's end effector.
This scenario is suitable for indirect teleoperation as presented in~\cite{fennelIntuitiveImmersiveTeleoperation2022}.
The resulting user-applied joint angles and torques in the excavator joint space are plotted in \reffig{fig:experiments:excavator}.
At time~$t_\text{a}$, the bucket joint reaches its upper limit, while the boom joint is already at its lower limit.
As a result, there is a sudden increase in the velocity of the stick joint.
This matches the expected behavior in reality and is a good example of why dynamically consistent rendering is important for a good user experience.
When~$t_\text{b}$ is reached, the arm of the excavator is fully extended, which is reflected by the stick joint being at its lower limit.
After~$t_\text{c}$ and~$t_\text{d}$, the rotation of the swing joint reaches its configured lower and upper limit, respectively, while the other joints can move freely.
Following time~$t_\text{e}$, the bucket is moved in a typical digging-and-unloading motion.

\subsection{Door}
\label{sec:door}

\begin{figure}
  \newcommand{\scaleDoor}{1.}
  \centering
  \subfigure[User wrench transformed into admittance joint space.]{
    \includegraphics[width=\scaleDoor\linewidth,trim={0 7.5mm 0 0},clip]{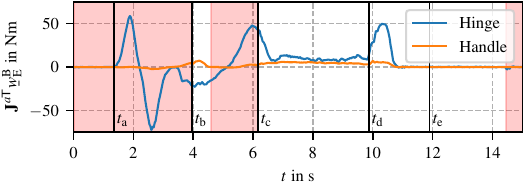}
    \label{fig:experiments:door:projected_user_wrench}
  }
  \subfigure[DT joint angles.]{
    \includegraphics[width=\scaleDoor\linewidth,trim={0 7.5mm 0 0},clip]{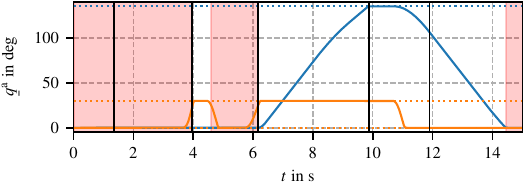}
    \label{fig:experiments:door:q_adm}
  }
  \subfigure[DT joint velocities.]{
    \includegraphics[width=\scaleDoor\linewidth]{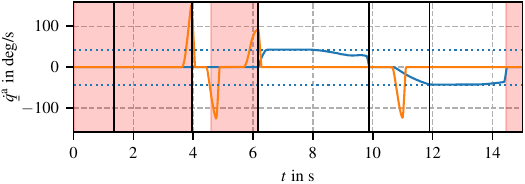}
    \label{fig:experiments:door:dq_adm}
  }
  \caption{Applied user torques and resulting DT motion for the door application. Limits are plotted as dotted lines. Highlighted areas indicate when the door is locked.}
  \label{fig:experiments:door}
\end{figure}

Following the concept of encountered-type haptics, HapticGiant is used to render the virtual door from \reffig{fig:experiments:scenarios:door}.
The door is modeled as a serial kinematic chain, with hinge and handle representing the joints.
The hinge is locked in place unless the handle joint exceeds a certain angle or until there is a small opening angle to mimic typical door behavior.
To implement this, the bounds \eqref{eq:control:rendering:task:admittance_limits_1} and \eqref{eq:control:rendering:task:admittance_limits_2} are modified accordingly at runtime. In addition, the case
\begin{equation}
  -\infty \le \alpha \le \infty, \quad \text{if locked}
\end{equation}
is added to \eqref{eq:control:rendering:task:constraint_torque_range:cases} to lock joints if necessary by activating the upper and lower limit simultaneously.
The springs that drive back the handle to its rest position and close the door are modeled by $\vek{\tau}{\adm}{\text{dri}}$ in \eqref{eq:control:rendering:task:dynamics}.

An exemplary user interaction is shown in \reffig{fig:experiments:door}.
After time~$t_\text{a}$, the user pushes and pulls the handle without pressing it down, which, as expected, results in no movement.
Then, the handle is pressed down at~$t_\text{b}$, but the door does not open, because the user is pulling instead of pushing.
In a second attempt, the user leans against the door, while pressing the handle, which causes the door to \enquote{pop} open at~$t_\text{c}$.
The door is then fully opened at~$t_\text{d}$ and released again.
Subsequently, the door slams shut while respecting the configured admittance velocity limit at~$t_\text{e}$, which can be interpreted as the behavior of a door closer with non-linear damping.

\subsection{User Studies}

\begin{figure}
    \centering
    \includegraphics[width=1.\linewidth]{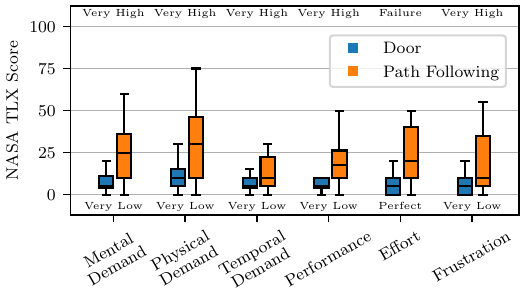}
    \caption{Box plots of NASA TLX scores for the door and path following tasks (\num{20} participants). Scores range from \num{0} to \num{100}, with lower scores being better.}
    \label{fig:experiments:nasatlx}
\end{figure}


While the HQP control scheme works objectively and consistently across users---applying the same optimization principles regardless of who is operating the system---the subjective experience and perceived usability may vary significantly between individuals.
The experiments presented thus far have been conducted by a single expert user and cannot be extrapolated to represent the general user population's experience. Therefore, we conducted preliminary user studies with \num{20} participants to evaluate the usability and immersion of the system from diverse end-user perspectives, providing insights into how effectively the technical capabilities of the system translate into practical usability for non-expert users. 
For this purpose we use the two tasks \emph{door} and \emph{path-following}.

The door task is based on the example in~\refsec{sec:door}. 
Each user is instructed to interact with the handle of HapticGiant as if it were a normal door handle, and is asked to open and walk through the rendered door.
The experiment is conducted without visual support in order to focus the experiment on the HapticGiant.
The second part of the study consists of following a path using the end effector. For that purpose, six checkpoints are marked on the ground, including the initial and final positions, which are distributed over the entire workspace. The user is instructed to use HapticGiant's end effector to design a path that connects these checkpoints. The checkpoints are positioned and aligned to force rotational movements and demonstrate the behavior at the Cartesian limits. 
Nevertheless, free user movement is made possible in this scenario.
Each of the experiments was repeated three times by each user. Users were then asked to assess the experience using the \emph{NASA Task Load Index} (NASA TLX).

The results are shown in~\reffig{fig:experiments:nasatlx}.
Overall, all median scores fall below the lower third of the scale, indicating intuitive and easy use of the system.
Notably, scores for the path-following task tend to be slightly larger than those for the door task, particularly in mental and physical demand. 
This can be explained by the different physical motions needed: opening a door is a highly intuitive action familiar to all participants, whereas following a path while encountering the Cartesian limits of HapticGiant requires spatial awareness and adaptation to system constraints. 
The door experiment leverages users' existing mental models, while path following demands more conscious attention to both the task requirements and the system's physical limitations. The low scores for the door experiment also show that the haptic representation of the digital twin provides a highly immersive and intuitive experience, demonstrating the capabilities of HapticGiant.

\subsection{Quantitative Evaluation}
Quantitative evaluation of kinesthetic haptic devices is difficult because there are many different experiments and quantities to choose from~\cite{samurPerformanceMetricsHaptic,seifiHaptipediaAcceleratingHaptic2019},
yet no single standard has been widely adopted.
%
%
%
Recently, the \emph{Haptify} benchmarking method~\cite{fazlollahiHaptifyMeasurementBasedBenchmarking2023} was introduced to establish a standardized framework for assessing grounded force-feedback devices.
%
%
Although Haptify primarily targets table-top haptic devices under impedance control with external instrumentation, most of its methods can be adapted to HapticGiant.

Using the Haptify benchmark, we obtain comparable metrics which include workspace shape, global free-space forces, global free-space vibrations, local dynamic forces and torques, frictionless surface rendering, and stiffness rendering. 
For further clarification of the mentioned metrics and the realization of the experiments see~\cite{fennelHighly2025}. 
The results are summarized in \reftab{tab:evaluation:quantitative:comparision_haptify} compared with two commercial table-top devices. 
From this data, we can see that HapticGiant has a workspace whose mean edge length that is about ten times longer than that of typical table-top devices.
At the same time, HapticGiant's mean free-space force is not considerably higher.
Moreover, the comparison also shows that HapticGiant can render stiff surfaces with higher accuracy and peak forces ten times higher than the commercial devices at the cost of higher planar forces.
Note that \SI{109}{\percent} stiffness rendering accuracy in the Haptify benchmark means that the stiffness is higher than expected. 
Further analysis of the measurement data revealed that the reason for this is the tracking error of the low-level joint controllers.
On the downside, HapticGiant has increased vibrations by a factor of ten with a lower spectral centroid, which limits the rendering of fine textures and surfaces.

\begin{table}[t]
    \centering
    \footnotesize
    \renewcommand*{\arraystretch}{1.2}
    \setlength{\tabcolsep}{2pt}
    \sisetup{text-series-to-math = true,propagate-math-font = true, exponent-product=\ensuremath{\cdot}}
        \begin{tabular}{llll}
            \toprule
            Metric                                                                    & HapticGiant        & Touch~\citeDatasheet{TouchTouchX} & Touch X~\citeDatasheet{TouchTouchX}
            \\ \midrule
            \makecell[l]{Workspace volume in \si{\cubic\cm}     }                                   & \textbf{\num{12.2e6}} & \num{11.8e3}              & \num{10.4e3}                \\
            \addlinespace[3pt]
            \makecell[l]{Mean free-space forces in \si{\newton} }                                   & \num{1.28}          & \num{0.98}              & \textbf{\num{0.87}}         \\
            \addlinespace[3pt]
            \makecell[l]{Free-space vibration \\ RMS$^\text{a}$ in \si{\meter\per\second\squared}} & \num{0.81}            & \num{0.07}                & \textbf{\num{0.04}}         \\
            \addlinespace[3pt]
            \makecell[l]{Free-space vibration \\ spectral centroid in \si{\hertz} }                    & \num{77.3}            & \num{123.0}               & \textbf{\num{162.8}}        \\
            \addlinespace[3pt]
            \makecell[l]{Frictionless surface \\ rendering forces in \si{\newton}  }                   & \num{1.96}            & \num{0.53}                & \textbf{\num{0.38}}         \\
            \addlinespace[3pt]
            \makecell[l]{Stiffness rendering \\ accuracy in \si{\percent}  }                           & \textbf{\num{109}}    & \num{47}                  & \num{80}                    \\
            \addlinespace[3pt]
            \makecell[l]{Stiffness peak force in \si{\newton} }                                     & \textbf{\num{76.7}}   & \num{2.7}                 & \num{5.7} 
            \\
            \bottomrule
            {\footnotesize $^\text{a}$Root Mean Square}
        \end{tabular}
    \caption[The Haptify metrics for HapticGiant and two commercial table-top devices.]{
        The Haptify metrics for HapticGiant and two commercial table-top devices.
        The values in the two columns on the right are taken from~\cite{fazlollahiHaptifyMeasurementBasedBenchmarking2023}.
        Bold text highlights the best value for each metric.
    }
    \label{tab:evaluation:quantitative:comparision_haptify}
\end{table}

\section{Conclusions and Outlook}
\label{sec:conclusion}

As demonstrated in the previous section, HapticGiant is a new and operational choice for adding kinesthetic haptic feedback to an application.
Especially the unprecedented size and dexterity of the workspace as well as the powerful high-level control scheme that intrinsically considers various limits, sets the system apart from competitors.
The experiment with the digital twin of the door is a perfect example of how this technology can be combined with head-mounted displays to create truly immersive AR/VR applications.
The rendering of the Cartesian admittance also revealed that the design goals \enquote{large workspace} and \enquote{optimized for the human arm} are fulfilled in practice.
Furthermore, it was shown that the HQP formulation of the control problem is a powerful approach to include all system constraints at once while still being real-time capable.
This also means that unskilled users can interact with HapticGiant without knowing too much about the system's limitations.
For example, they might not even notice an active constraint in case of approaching the manipulator's singularity, because they can still move a dynamically consistent DT.
But even if they do, the system will be kept within its safe operating limits.

In future work, we plan to enhance different parts of the system.
In particular, optimizing the low-level joint tracking controllers will improve haptic quality, especially when effects such as non-linear friction and finite link stiffness are taken into account.
The hierarchical controller formulation can be used to implement further tasks.
For example, we consider integrating the PPU optimization as a task.
To increase functional safety, a compliance task to mitigate the impact of unintended human-robot collisions seems to be an interesting extension.
On the application side, the utilization of HapticGiant for direct teleoperation will be studied.
Moreover, combining the intention recognition from~\cite{fennelIntentionEstimationRecurrent} with a suitable motion planning algorithm will lead to interesting research and application opportunities in AR/VR environments.

\section*{Acknowledgments}
This work would not have been possible without the dedication and joint effort of many people, including
Philipp Bl\"attner,
Giulio Broghammer,
Lukas Driller,
Sascha Faber,
Csaba Freiberger,
Serge Garbay,
Stefan Geyer,
Jakob Kessler,
Achim Langend\"orfer,
Felix Pfaff,
Alexander Riffel,
Marcel Stahl,
Jakob Wandel, and
Antonio Zea.

\bibliographystyle{bib/IEEEtran}
\bibliography{bib/IEEEfull,bib/literature,bib/datasheets}

\bibliographystyleDatasheet{bib/IEEEtran}
\bibliographyDatasheet{bib/IEEEfull,bib/datasheets}





\begin{IEEEbiography}[{\includegraphics[width=1in,height=1.25in,clip,keepaspectratio]{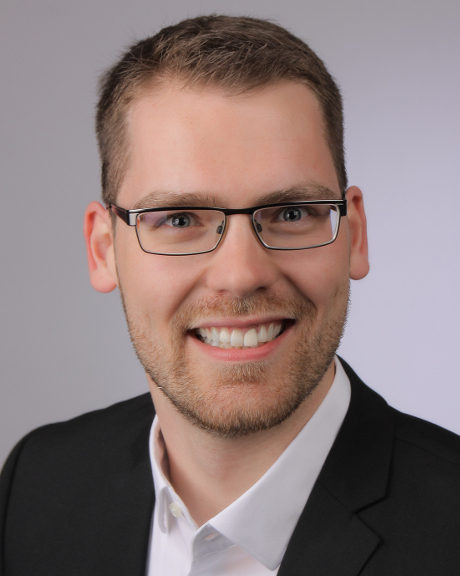}}]{Michael Fennel}
  is a researcher and PhD student at the Intelligent Sensor-Actuator-Systems Laboratory of the Karlsruhe Institute of Technology, where he also received his Bachelor's degree in 2016 and his Master's degree in 2020, both in electrical engineering.
  His research interests include haptic systems, robotics, human-machine interaction, unmanned aircraft systems, and autonomous systems.
\end{IEEEbiography}

\begin{IEEEbiography}[{\includegraphics[width=1in,height=1.25in,clip,keepaspectratio]{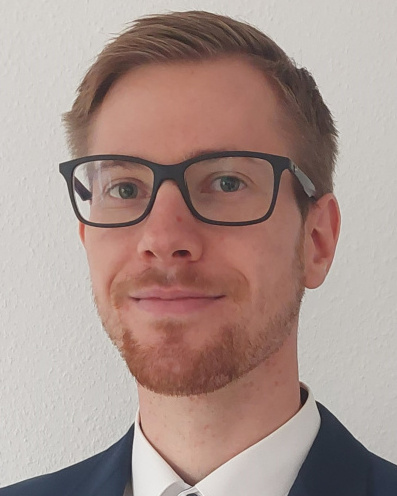}}]{Markus Walker}
  received his Master's degree in Electrical Engineering and Information Technology in 2022 at the Karlsruhe Institute of Technology.
  Since then, he has been a researcher and PhD student at the Intelligent Sensor-Actuator-Systems Laboratory at Karlsruhe Institute of Technology.
  He is interested in Bayesian neural networks, non-linear systems, and optimal control.
\end{IEEEbiography}

\begin{IEEEbiography}[{\includegraphics[width=1in,height=1.25in,clip,keepaspectratio]{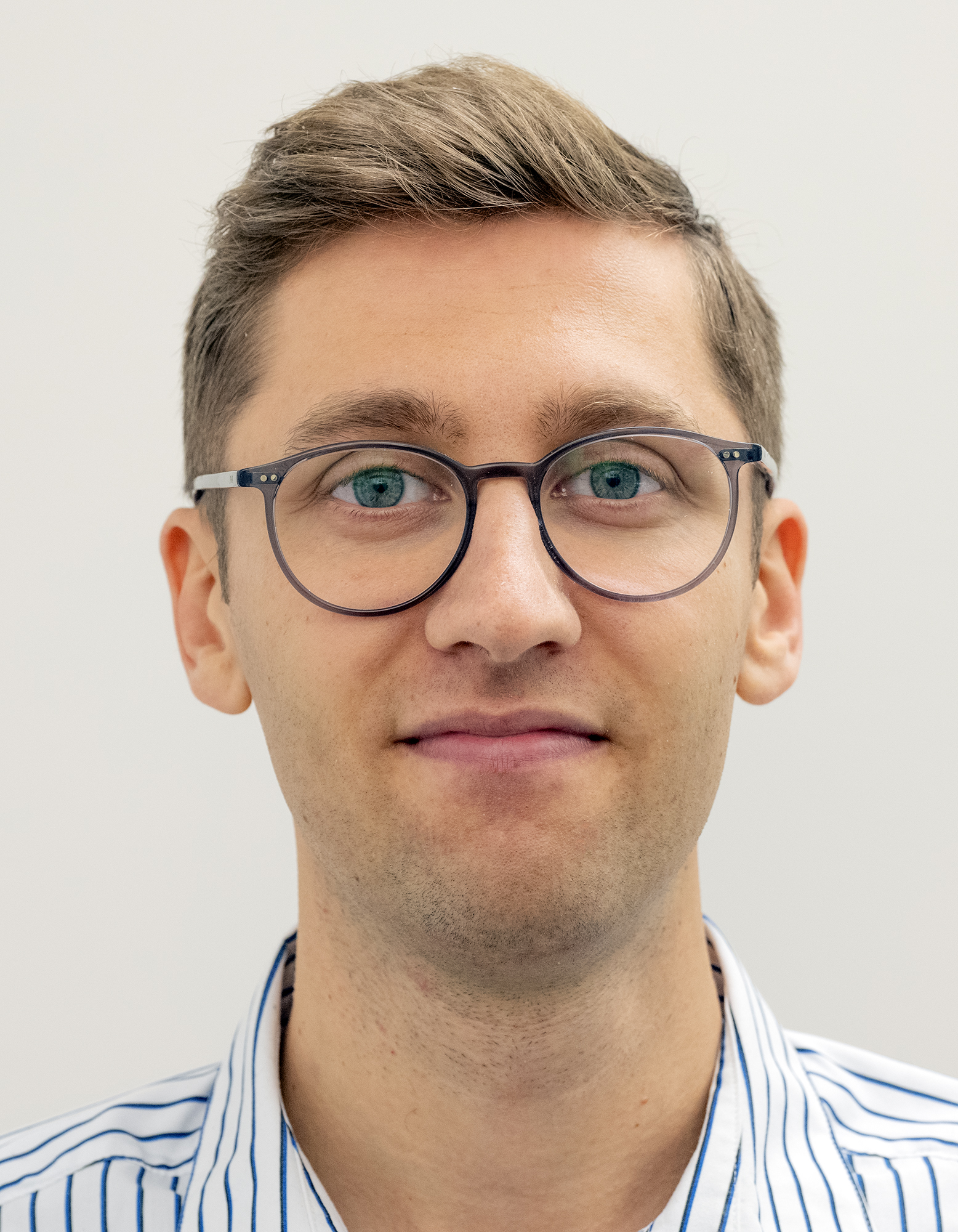}}]{Dominik Pikos}
  received his Master`s degree in product and system development in 2023 at the Technical University of Applied Sciences W\"urzburg-Schweinfurt. Since 2024 he has been a PhD student at the Intelligent Sensor-Actuator-Systems Laboratory at Karlsruhe Institute of Technology. His research focuses on highly immersive telepresence with mixed reality and haptic interfaces. 
\end{IEEEbiography}

\begin{IEEEbiography}[{\includegraphics[width=1in,height=1.25in,clip,keepaspectratio]{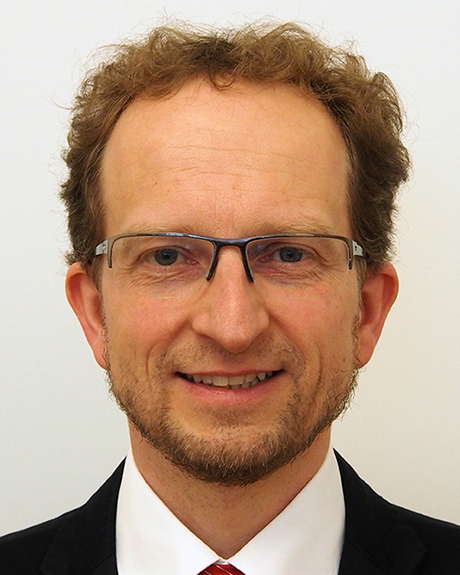}}]{Uwe D. Hanebeck}
  is a chaired professor at the Intelligent Sensor-Actuator-Systems Laboratory of the Karlsruhe Institute of Technology (KIT).
  He obtained his PhD degree in 1997 and his habilitation degree in 2003.
  His research interests are in the areas of information fusion, non-linear state estimation, stochastic modeling, system identification, and control.
  He is author and co-author of more than 600 publications in various high-ranking journals and conferences and an IEEE Fellow.
\end{IEEEbiography}



\vfill

\end{document}

%% file: macros.tex
\usepackage{graphicx}
\usepackage[raggedright]{subfigure}
\usepackage{xcolor}

\usepackage{amsmath,amssymb,amsfonts}
\usepackage{accents}
\usepackage{nicefrac}
\usepackage{mathtools}

\usepackage{import}
\usepackage{siunitx}
\usepackage{cite}
\usepackage{textcomp}
\usepackage{blindtext}
\usepackage{xurl}
\usepackage{url}
\usepackage{csquotes}
\usepackage[inline]{enumitem}
\usepackage{stfloats}

\usepackage{algorithmicx}
\usepackage[noend]{algpseudocode}
\usepackage{algorithm}

\usepackage{multirow}
\usepackage{booktabs}

\usepackage{tikz}
\usetikzlibrary{positioning,calc,fit,shapes.geometric, decorations.pathmorphing, decorations.pathreplacing, decorations.shapes,}

\usepackage{caption}
\usepackage{multibib}
\newcites{Datasheet}{Datasheets}

\usepackage{xparse}
\NewDocumentCommand\vek{mgg}{
  \ensuremath{
    \underaccent{\bar}{#1}\IfNoValueTF{#2}{}{^{#2}}\IfNoValueTF{#3}{}{_{#3}}
  }
}

\NewDocumentCommand\mat{mgg}{
  \ensuremath{
    \mathbf{#1}\IfNoValueTF{#2}{}{^{#2}}\IfNoValueTF{#3}{}{_{#3}}
  }
}

\NewDocumentCommand\q{mg}{
  \ensuremath{
    \underaccent{\bar}{q}^{#1}\IfNoValueTF{#2}{}{_{#2}}
  }
}
\NewDocumentCommand\dq{mg}{
  \ensuremath{
    \underaccent{\bar}{\dot{q}}^{#1}\IfNoValueTF{#2}{}{_{#2}}
  }
}
\NewDocumentCommand\ddq{mg}{
  \ensuremath{
    \underaccent{\bar}{\ddot{q}}^{#1}\IfNoValueTF{#2}{}{_{#2}}
  }
}

\newcommand{\adm}{\text{a}} 
\newcommand{\manip}{\text{m}} 
\newcommand{\ppu}{\text{p}} 
\newcommand{\sing}{\text{sing}} 
\newcommand{\ee}{\text{ee}} 
\newcommand{\elbow}{\text{elbow}} 

\newcommand{\fdtee}{\text{E}}
\newcommand{\fhree}{\text{H}}

\newcommand{\fhrbase}{\text{B}}

\newcommand{\mathspace}{\,}
\newcommand{\norm}[1]{\left\lVert#1\right\rVert}
\newcommand{\tranT}{\mkern-1.5mu\mathsf{T}}
\newcommand{\tran}{^{\tranT}}

\newcommand{\reffig}[1]{Fig.~\ref{#1}}
\newcommand{\reftab}[1]{Table~\ref{#1}}

\newcommand{\refsec}[1]{Section~\ref{#1}}

%% file: figures/generated/definitions.pdf_tex
\begingroup%
  \makeatletter%
  \providecommand\color[2][]{%
    \errmessage{(Inkscape) Color is used for the text in Inkscape, but the package 'color.sty' is not loaded}%
    \renewcommand\color[2][]{}%
  }%
  \providecommand\transparent[1]{%
    \errmessage{(Inkscape) Transparency is used (non-zero) for the text in Inkscape, but the package 'transparent.sty' is not loaded}%
    \renewcommand\transparent[1]{}%
  }%
  \providecommand\rotatebox[2]{#2}%
  \newcommand*\fsize{\dimexpr\f@size pt\relax}%
  \newcommand*\lineheight[1]{\fontsize{\fsize}{#1\fsize}\selectfont}%
  \ifx\svgwidth\undefined%
    \setlength{\unitlength}{181.35507058bp}%
    \ifx\svgscale\undefined%
      \relax%
    \else%
      \setlength{\unitlength}{\unitlength * \real{\svgscale}}%
    \fi%
  \else%
    \setlength{\unitlength}{\svgwidth}%
  \fi%
  \global\let\svgwidth\undefined%
  \global\let\svgscale\undefined%
  \makeatother%
  \begin{picture}(1,0.88311554)%
    \lineheight{1}%
    \setlength\tabcolsep{0pt}%
    \put(0,0){\includegraphics[width=\unitlength,page=1]{definitions.pdf}}%
    \put(0.54468164,0.59292954){\color[rgb]{0,0,0}\makebox(0,0)[t]{\lineheight{1.25}\smash{\begin{tabular}[t]{c}P\end{tabular}}}}%
    \put(0.44386677,0.13824185){\color[rgb]{0,0,0}\makebox(0,0)[t]{\lineheight{1.25}\smash{\begin{tabular}[t]{c}H\end{tabular}}}}%
    \put(0.76520623,0.70383482){\color[rgb]{0,0.55294118,0.14117647}\makebox(0,0)[lt]{\lineheight{1.25}\smash{\begin{tabular}[t]{l}\qppu{1}\end{tabular}}}}%
    \put(0.4757243,0.55116313){\color[rgb]{0.80784314,0.02352941,0.02352941}\makebox(0,0)[rt]{\lineheight{1.25}\smash{\begin{tabular}[t]{r}$x$\end{tabular}}}}%
    \put(0.53239614,0.50383014){\color[rgb]{0,0.77647059,0}\makebox(0,0)[lt]{\lineheight{1.25}\smash{\begin{tabular}[t]{l}$y$\end{tabular}}}}%
    \put(0.60746054,0.62785102){\color[rgb]{0,0,0.86666667}\makebox(0,0)[lt]{\lineheight{1.25}\smash{\begin{tabular}[t]{l}$z$\end{tabular}}}}%
    \put(0.83404907,0.74155183){\color[rgb]{0,0,0}\makebox(0,0)[t]{\lineheight{1.25}\smash{\begin{tabular}[t]{c}B\end{tabular}}}}%
    \put(0.37374957,0.18733954){\color[rgb]{0,0,0}\makebox(0,0)[t]{\lineheight{1.25}\smash{\begin{tabular}[t]{c}E\end{tabular}}}}%
    \put(0.16650993,0.08658164){\color[rgb]{0,0,0}\makebox(0,0)[t]{\lineheight{1.25}\smash{\begin{tabular}[t]{c}R\end{tabular}}}}%
    \put(0.62617826,0.67674623){\color[rgb]{0,0.55294118,0.14117647}\makebox(0,0)[lt]{\lineheight{1.25}\smash{\begin{tabular}[t]{l}\qppu{2}\end{tabular}}}}%
    \put(0.59560155,0.58552923){\color[rgb]{0,0.52156863,0.84313725}\makebox(0,0)[lt]{\lineheight{1.25}\smash{\begin{tabular}[t]{l}\qmanip{1}\end{tabular}}}}%
    \put(0.32076541,0.60837127){\color[rgb]{0,0.52156863,0.84313725}\makebox(0,0)[lt]{\lineheight{1.25}\smash{\begin{tabular}[t]{l}\qmanip{2}\end{tabular}}}}%
    \put(0.08347925,0.57162332){\color[rgb]{0,0.52156863,0.84313725}\makebox(0,0)[lt]{\lineheight{1.25}\smash{\begin{tabular}[t]{l}\qmanip{3}\end{tabular}}}}%
    \put(0.01825457,0.26714242){\color[rgb]{0,0.52156863,0.84313725}\makebox(0,0)[lt]{\lineheight{1.25}\smash{\begin{tabular}[t]{l}\qmanip{4}\end{tabular}}}}%
    \put(0.28506751,0.23354434){\color[rgb]{0,0.52156863,0.84313725}\makebox(0,0)[lt]{\lineheight{1.25}\smash{\begin{tabular}[t]{l}\qmanip{5}\end{tabular}}}}%
    \put(0.28191771,0.12435096){\color[rgb]{0,0.52156863,0.84313725}\makebox(0,0)[lt]{\lineheight{1.25}\smash{\begin{tabular}[t]{l}\qmanip{6}\end{tabular}}}}%
    \put(0.17337708,0.14750132){\color[rgb]{0.80784314,0.02352941,0.02352941}\makebox(0,0)[rt]{\lineheight{1.25}\smash{\begin{tabular}[t]{r}\qadm{1}\end{tabular}}}}%
    \put(0.19502261,0.31465028){\color[rgb]{0.80784314,0.02352941,0.02352941}\makebox(0,0)[lt]{\lineheight{1.25}\smash{\begin{tabular}[t]{l}\qadm{2}\end{tabular}}}}%
    \put(0.30500506,0.3850755){\color[rgb]{0.80784314,0.02352941,0.02352941}\makebox(0,0)[lt]{\lineheight{1.25}\smash{\begin{tabular}[t]{l}\qadm{n-1}\end{tabular}}}}%
    \put(0.40586956,0.3191049){\color[rgb]{0.80784314,0.02352941,0.02352941}\makebox(0,0)[lt]{\lineheight{1.25}\smash{\begin{tabular}[t]{l}\qadm{n}\end{tabular}}}}%
  \end{picture}%
\endgroup%

%% file: figures/generated/architecture_overview.pdf_tex
\begingroup%
  \makeatletter%
  \providecommand\color[2][]{%
    \errmessage{(Inkscape) Color is used for the text in Inkscape, but the package 'color.sty' is not loaded}%
    \renewcommand\color[2][]{}%
  }%
  \providecommand\transparent[1]{%
    \errmessage{(Inkscape) Transparency is used (non-zero) for the text in Inkscape, but the package 'transparent.sty' is not loaded}%
    \renewcommand\transparent[1]{}%
  }%
  \providecommand\rotatebox[2]{#2}%
  \newcommand*\fsize{\dimexpr\f@size pt\relax}%
  \newcommand*\lineheight[1]{\fontsize{\fsize}{#1\fsize}\selectfont}%
  \ifx\svgwidth\undefined%
    \setlength{\unitlength}{252.00000433bp}%
    \ifx\svgscale\undefined%
      \relax%
    \else%
      \setlength{\unitlength}{\unitlength * \real{\svgscale}}%
    \fi%
  \else%
    \setlength{\unitlength}{\svgwidth}%
  \fi%
  \global\let\svgwidth\undefined%
  \global\let\svgscale\undefined%
  \makeatother%
  \begin{picture}(1,0.49493812)%
    \lineheight{1}%
    \setlength\tabcolsep{0pt}%
    \put(0,0){\includegraphics[width=\unitlength,page=1]{architecture_overview.pdf}}%
    \put(0.34870641,0.1799775){\color[rgb]{0,0,0}\makebox(0,0)[lt]{\lineheight{1.04999995}\smash{\begin{tabular}[t]{l}Real-Time\end{tabular}}}}%
    \put(0,0){\includegraphics[width=\unitlength,page=2]{architecture_overview.pdf}}%
    \put(0.65122786,0.25022601){\color[rgb]{0,0,0}\makebox(0,0)[t]{\lineheight{1.04999995}\smash{\begin{tabular}[t]{c}Safety\end{tabular}}}}%
    \put(0,0){\includegraphics[width=\unitlength,page=3]{architecture_overview.pdf}}%
    \put(0.809519,0.44338905){\color[rgb]{0,0,0}\makebox(0,0)[t]{\lineheight{1.04999995}\smash{\begin{tabular}[t]{c}State\\Estimation\end{tabular}}}}%
    \put(0,0){\includegraphics[width=\unitlength,page=4]{architecture_overview.pdf}}%
    \put(0.51665543,0.33835079){\color[rgb]{0,0,0}\makebox(0,0)[t]{\lineheight{1.04999995}\smash{\begin{tabular}[t]{c}Simulation\end{tabular}}}}%
    \put(0,0){\includegraphics[width=\unitlength,page=5]{architecture_overview.pdf}}%
    \put(0.52108386,0.44314492){\color[rgb]{0,0,0}\makebox(0,0)[t]{\lineheight{1.04999995}\smash{\begin{tabular}[t]{c}Hardware\\Interface\end{tabular}}}}%
    \put(0,0){\includegraphics[width=\unitlength,page=6]{architecture_overview.pdf}}%
    \put(0.80911882,0.33835079){\color[rgb]{0,0,0}\makebox(0,0)[t]{\lineheight{1.04999995}\smash{\begin{tabular}[t]{c}Controller\end{tabular}}}}%
    \put(0,0){\includegraphics[width=\unitlength,page=7]{architecture_overview.pdf}}%
    \put(0.64071527,0.09301343){\color[rgb]{0,0,0}\makebox(0,0)[t]{\lineheight{1.04999995}\smash{\begin{tabular}[t]{c}Logging\end{tabular}}}}%
    \put(0,0){\includegraphics[width=\unitlength,page=8]{architecture_overview.pdf}}%
    \put(0.83206819,0.09975215){\color[rgb]{0,0,0}\makebox(0,0)[lt]{\lineheight{1.04999995}\smash{\begin{tabular}[t]{l}...\end{tabular}}}}%
    \put(0.34870641,0.02249719){\color[rgb]{0,0,0}\makebox(0,0)[lt]{\lineheight{1.04999995}\smash{\begin{tabular}[t]{l}Non Real-Time\end{tabular}}}}%
    \put(0,0){\includegraphics[width=\unitlength,page=9]{architecture_overview.pdf}}%
    \put(0.16873568,0.18100612){\color[rgb]{0,0,0}\makebox(0,0)[t]{\lineheight{1.04999995}\smash{\begin{tabular}[t]{c}iviz\end{tabular}}}}%
    \put(0,0){\includegraphics[width=\unitlength,page=10]{architecture_overview.pdf}}%
    \put(0.15177544,0.04764467){\color[rgb]{0,0,0}\makebox(0,0)[t]{\lineheight{1.04999995}\smash{\begin{tabular}[t]{c}3rd Party Software\end{tabular}}}}%
    \put(0,0){\includegraphics[width=\unitlength,page=11]{architecture_overview.pdf}}%
    \put(0.4383154,0.1089084){\color[rgb]{0,0,0}\makebox(0,0)[t]{\lineheight{1.04999995}\smash{\begin{tabular}[t]{c}Intention\\Recognition\end{tabular}}}}%
  \end{picture}%
\endgroup%

%% file: figures/generated/controller.pdf_tex
\begingroup%
  \makeatletter%
  \providecommand\color[2][]{%
    \errmessage{(Inkscape) Color is used for the text in Inkscape, but the package 'color.sty' is not loaded}%
    \renewcommand\color[2][]{}%
  }%
  \providecommand\transparent[1]{%
    \errmessage{(Inkscape) Transparency is used (non-zero) for the text in Inkscape, but the package 'transparent.sty' is not loaded}%
    \renewcommand\transparent[1]{}%
  }%
  \providecommand\rotatebox[2]{#2}%
  \newcommand*\fsize{\dimexpr\f@size pt\relax}%
  \newcommand*\lineheight[1]{\fontsize{\fsize}{#1\fsize}\selectfont}%
  \ifx\svgwidth\undefined%
    \setlength{\unitlength}{515.90551181bp}%
    \ifx\svgscale\undefined%
      \relax%
    \else%
      \setlength{\unitlength}{\unitlength * \real{\svgscale}}%
    \fi%
  \else%
    \setlength{\unitlength}{\svgwidth}%
  \fi%
  \global\let\svgwidth\undefined%
  \global\let\svgscale\undefined%
  \makeatother%
  \begin{picture}(1,0.21978022)%
    \lineheight{1}%
    \setlength\tabcolsep{0pt}%
    \put(0,0){\includegraphics[width=\unitlength,page=1]{controller.pdf}}%
    \put(0.63116372,0.15504367){\makebox(0,0)[t]{\lineheight{1.25}\smash{\begin{tabular}[t]{c}Visualization\end{tabular}}}}%
    \put(0.88475629,0.08741899){\makebox(0,0)[t]{\lineheight{1.25}\smash{\begin{tabular}[t]{c}Manipulator\end{tabular}}}}%
    \put(0.76641309,0.15504367){\makebox(0,0)[t]{\lineheight{1.25}\smash{\begin{tabular}[t]{c}\Je\end{tabular}}}}%
    \put(0.75796,0.10432516){\makebox(0,0)[t]{\lineheight{1.25}\smash{\begin{tabular}[t]{c}$-$\end{tabular}}}}%
    \put(0.77768387,0.10150747){\makebox(0,0)[lt]{\lineheight{1.25}\smash{\begin{tabular}[t]{l}\taucmd\end{tabular}}}}%
    \put(0.31839956,0.16913215){\makebox(0,0)[lt]{\lineheight{1.25}\smash{\begin{tabular}[t]{l}\ddqa\end{tabular}}}}%
    \put(0.4085658,0.16913215){\makebox(0,0)[lt]{\lineheight{1.25}\smash{\begin{tabular}[t]{l}\dqa\end{tabular}}}}%
    \put(0.49873205,0.16913215){\makebox(0,0)[lt]{\lineheight{1.25}\smash{\begin{tabular}[t]{l}\qa\end{tabular}}}}%
    \put(0.31839956,0.10150749){\makebox(0,0)[lt]{\lineheight{1.25}\smash{\begin{tabular}[t]{l}\ddqm\end{tabular}}}}%
    \put(0.4085658,0.10150749){\makebox(0,0)[lt]{\lineheight{1.25}\smash{\begin{tabular}[t]{l}\dqm\end{tabular}}}}%
    \put(0.49873205,0.10150749){\makebox(0,0)[lt]{\lineheight{1.25}\smash{\begin{tabular}[t]{l}\qm\end{tabular}}}}%
    \put(0.77768387,0.03388278){\makebox(0,0)[lt]{\lineheight{1.25}\smash{\begin{tabular}[t]{l}\vppucmd\end{tabular}}}}%
    \put(0.88475629,0.01979431){\makebox(0,0)[t]{\lineheight{1.25}\smash{\begin{tabular}[t]{c}PPU\end{tabular}}}}%
    \put(0.25903381,0.12108123){\makebox(0,0)[t]{\lineheight{1.25}\smash{\begin{tabular}[t]{c}HQP\end{tabular}}}}%
    \put(0.38320654,0.15504367){\makebox(0,0)[t]{\lineheight{1.25}\smash{\begin{tabular}[t]{c}$\int$\end{tabular}}}}%
    \put(0.38320654,0.08741899){\makebox(0,0)[t]{\lineheight{1.25}\smash{\begin{tabular}[t]{c}$\int$\end{tabular}}}}%
    \put(0.47337279,0.08741899){\makebox(0,0)[t]{\lineheight{1.25}\smash{\begin{tabular}[t]{c}$\int$\end{tabular}}}}%
    \put(0.47337279,0.15504367){\makebox(0,0)[t]{\lineheight{1.25}\smash{\begin{tabular}[t]{c}$\int$\end{tabular}}}}%
    \put(0.63116372,0.08741899){\makebox(0,0)[t]{\lineheight{1.25}\smash{\begin{tabular}[t]{c}Joint Controller\end{tabular}}}}%
    \put(0.63116372,0.01979431){\makebox(0,0)[t]{\lineheight{1.25}\smash{\begin{tabular}[t]{c}PPU Optimization\end{tabular}}}}%
    \put(0.16342632,0.1034134){\makebox(0,0)[t]{\lineheight{1.25}\smash{\begin{tabular}[t]{c}$-$\end{tabular}}}}%
    \put(0.1155255,0.12123134){\makebox(0,0)[lt]{\lineheight{1.25}\smash{\begin{tabular}[t]{l}\wmeas\end{tabular}}}}%
    \put(0.12397859,0.05360665){\makebox(0,0)[lt]{\lineheight{1.25}\smash{\begin{tabular}[t]{l}\ddqppu\end{tabular}}}}%
    \put(0.12397859,0.03106509){\makebox(0,0)[lt]{\lineheight{1.25}\smash{\begin{tabular}[t]{l}\dqppu\end{tabular}}}}%
    \put(0.12397859,0.00852353){\makebox(0,0)[lt]{\lineheight{1.25}\smash{\begin{tabular}[t]{l}\qppu\end{tabular}}}}%
    \put(0.1746971,0.07896591){\makebox(0,0)[t]{\lineheight{1.25}\smash{\begin{tabular}[t]{c}\wcmd\end{tabular}}}}%
    \put(0,0){\includegraphics[width=\unitlength,page=2]{controller.pdf}}%
  \end{picture}%
\endgroup%